\pdfoutput=1
\documentclass[10pt,twocolumn,letterpaper]{article}

\usepackage{iccv}
\usepackage{times}
\usepackage{epsfig}
\usepackage{graphicx}
\usepackage{amsmath}
\usepackage{amssymb}
\usepackage{booktabs}
\usepackage{multirow}
\usepackage[dvipsnames]{xcolor}
\usepackage{color, colortbl}
\usepackage{subcaption}
\usepackage[section]{placeins}
\usepackage{url}

\usepackage{marvosym}
\usepackage[accsupp]{axessibility}

\definecolor{Gray}{gray}{0.9}
\newcommand{\cg}{\cellcolor{Gray}}

\usepackage[pagebackref=true,breaklinks=true,letterpaper=true,colorlinks,bookmarks=false]{hyperref}

\iccvfinalcopy

\begin{document}

\title{Semantically Coherent Out-of-Distribution Detection}

\author{Jingkang Yang$^{1}$, Haoqi Wang$^{2}$, Litong Feng$^{2}$, Xiaopeng Yan$^{2}$, Huabin Zheng$^{2}$,  \\Wayne Zhang$^{2,3,4}$, Ziwei Liu$^{1}$\textsuperscript{\Letter}\\
$^{1}$~S-Lab, Nanyang Technological University \qquad
$^{2}$~SenseTime Research\\
$^{3}$~Qing Yuan Research Institute, Shanghai Jiao Tong University\\
$^{4}$~Shanghai AI Laboratory, Shanghai, China\\
{\tt\small 
jingkang001@ntu.edu.sg, lastname.firstname@sensetime.com, ziwei.liu@ntu.edu.sg}
}

\maketitle

\begin{abstract}
Current out-of-distribution~(OOD) detection benchmarks are commonly built by defining one dataset as in-distribution~(ID) and all others as OOD. 
However, these benchmarks unfortunately introduce some unwanted and impractical goals, \eg.,~to perfectly distinguish CIFAR dogs from ImageNet dogs, even though they have the same semantics and negligible covariate shifts.
These unrealistic goals will result in an extremely narrow range of model capabilities, greatly limiting their use in real applications.
To overcome these drawbacks, we re-design the benchmarks and propose the \textbf{semantically coherent out-of-distribution detection (SC-OOD)}.
On the SC-OOD benchmarks, existing methods suffer from large performance degradation, suggesting that they are extremely sensitive to low-level discrepancy between data sources while ignoring their inherent semantics.
To develop an effective SC-OOD detection approach, 
we leverage an external unlabeled set and design a concise framework featured by \textbf{unsupervised dual grouping (UDG)} for the joint modeling of ID and OOD data.
The proposed UDG can not only enrich the semantic knowledge of the model by exploiting unlabeled data in an unsupervised manner, but also distinguish ID/OOD samples to enhance ID classification and OOD detection tasks simultaneously.
Extensive experiments demonstrate that our approach achieves the state-of-the-art performance on SC-OOD benchmarks. 
Code and benchmarks are provided on our project page: \href{https://jingkang50.github.io/projects/scood}{https://jingkang50.github.io/projects/scood}.

\end{abstract}
\section{Introduction}
\label{S:intro}

\begin{figure}[t]
\begin{center}
\includegraphics[width=\linewidth]{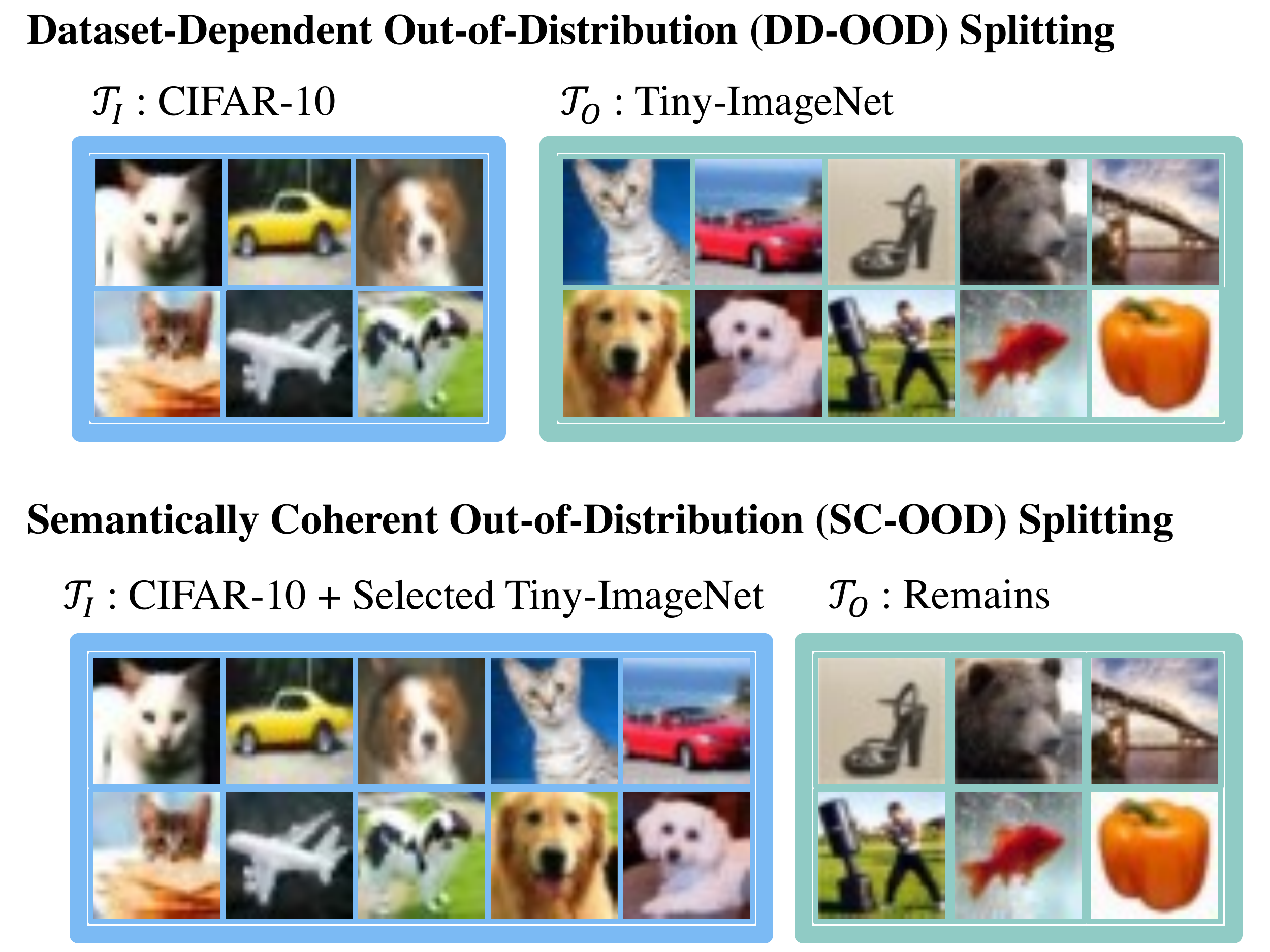}
\end{center}
\caption{\textbf{Dataset-dependent OOD (DD-OOD) \vs. Semantically Coherent OOD (SC-OOD).} We notice that a certain number of DD-OOD samples have the same semantics as ID, with negligible covariate shifts. Here, we redirect these samples back to ID in the SC-OOD setting.}
\label{fig:problem_statement}
\end{figure}

Although dominated on visual recognition~\cite{dnn,imagenet}, 
deep learning models are still notorious for the following two drawbacks:
\textbf{1)} their performance endures a significant drop when test data distribution has a large covariate shift from training~\cite{dgsurvey};
\textbf{2)} they tend to recklessly classify a test image into a certain training class, even though it has a semantic shift from training, that is, it may not belong to any training class~\cite{baseline}.
Those defects seriously reduce the model trustworthiness and hinder their deployment in real, especially high-risk applications~\cite{trustsurvey,autodriving}.
To solve the problem, out-of-distribution (OOD) detection aims to distinguish and reject test samples with either covariate shifts or semantic shifts or both, so as to prevent models trained on in-distribution (ID) data from producing unreliable predictions~\cite{baseline}.

Existing OOD detection methods mostly focus on calibrating the distribution of the softmax layer~\cite{baseline} through 
temperature scaling~\cite{odin}, generative models~\cite{mahalanobis,S},
or ensemble methods~\cite{waic,eloc}.
Other solutions collect enormous OOD samples to learn the ID/OOD discrepancy during training~\cite{objectosphere,hendrycks18oe,mcd}.
Appealing experimental results are achieved by existing methods. 
For example, MCD~\cite{mcd} reports \emph{near-perfect} scores across the classic benchmarks.
The OOD detection problem seems completely solved.

However, by scrutinizing the common-used OOD detection benchmarks~\cite{baseline,odin,hendrycks18oe},
we discover some irrationality on OOD splitting.
Under the assumption that different datasets represent different data distributions, current benchmarks are commonly built by defining one dataset as ID and all others as OOD.
Figure~\ref{fig:problem_statement}\textcolor{red}{-a} shows one popular benchmark that uses the entire CIFAR-10 test data as ID and the entire Tiny-ImageNet test data as OOD.
However, we observed that around 15\% Tiny-ImageNet test samples actually shares the same semantics with CIFAR-10's ID categories (\textit{ref.}~Section~\ref{S:benchmark}).
For example, Tiny-ImageNet contains six dog-breeds~(\eg golden retriever, Chihuahua) that match CIFAR-10's class of dog, while their covariate shifts are negligible.
In this case, the perfect performance on the above dataset-dependent OOD~(DD-OOD) benchmarks may indicate that models are attempting to overfit the low-level discrepancy on the negligible covariate shifts between data sources while ignoring inherent semantics. 
This fails to meet the requirement of realistic model deployment.

To overcome the drawbacks of the DD-OOD benchmarks, in this work,
we re-design \textbf{semantically coherent out-of-distribution detection~(SC-OOD)} benchmarks, which re-organize ID/OOD set based on semantics and only focus on real images where the covariate shift can be ignored, as depicted in Figure~\ref{fig:problem_statement}\textcolor{red}{-b}.
In this case, the ID set becomes semantically coherent and different from OOD.
Existing methods suffer a large performance degradation on revised SC-OOD benchmarks as shown in Table~\ref{T:DDOOD},
indicating that the OOD detection problem is still unresolved.
~\footnote{In Table~\ref{T:DDOOD},
both DD and SC-OOD benchmarks consider CIFAR-10 as ID and Tiny-ImageNet as OOD.
All methods are tested using their released DenseNet models.
AUPR corresponds to AUPR-Out in Section~\ref{S:benchmark}.}

\begin{table}[t]
\setlength{\tabcolsep}{3.7pt}
\centering
\caption{\textbf{Performance of ODIN~\cite{odin}, MCD~\cite{mcd}, and Energy-Based OOD~(EBO)~\cite{energyood} on DD-OOD/SC-OOD settings.}
Previous methods achieve nearly perfect results on DD-OOD but suffer a drastic drop on SC-OOD.\textcolor{red}{~$^1$}}
\label{T:DDOOD}
\begin{tabular}{l|c|c||c|c||c|c}
\toprule
\multirow{2}{*}{Method} & \multicolumn{2}{c||}{FPR95~$\downarrow$} & \multicolumn{2}{c||}{AUROC~$\uparrow$} & \multicolumn{2}{c}{AUPR~$\uparrow$}     \\ \cmidrule(l){2-7} 
                & DD     & \cg SC      & DD      & \cg SC     & DD       & \cg SC        \\ \midrule
ODIN~(ICLR18)   & 0.46   & \cg 55.0    & 99.8    & \cg 88.8   & 99.8     & \cg 84.2      \\
EBO~(NeurIPS20) & 1.56   & \cg 50.6    & 99.5    & \cg 90.4   & 99.4     & \cg 85.4       \\
MCD~(ICCV19)    & 0.01   & \cg 68.6    & 99.9    & \cg 88.9   & 99.9     & \cg 82.1      \\
\bottomrule
\end{tabular}
\end{table}

For an effective SC-OOD approach, we leverage an external unlabeled set like OE~\cite{hendrycks18oe}. \textit{Different from OE~\cite{hendrycks18oe} whose unlabeled set is purely OOD, our unlabeled set is contaminated by a portion of ID samples.}
We believe it is a more realistic setting, as a powerful image crawler can easily prepare millions of unlabeled data but will inevitably introduce ID samples that are expensive to be purified.

With a realistic unlabeled set for SC-OOD, we design an elegant framework featured by \textbf{unsupervised dual grouping~(UDG)} for the joint modeling of labeled and unlabeled data.
The proposed UDG enhances the semantic expression ability of the model by exploring unlabeled data with an unsupervised deep clustering task, and the grouping information generated by the auxiliary task can also dynamically separate the ID and OOD samples in the unlabeled set.
ID samples separated from the unlabeled set will join other given ID samples for classifier training, and the rest will be forced to produce a uniform posterior distribution like other OE methods~\cite{hendrycks18oe}.
In this way, ID classification and OOD detection performances are simultaneously improved.

To summarize, the contributions of our paper are:
\textbf{1)} We highlight the problem of current OOD detection benchmarks and re-design them to address semantic coherency in out-of-distribution detection.
\textbf{2)} A concise framework using realistic unlabeled data is proposed, featuring the unsupervised dual grouping which not only enriches the semantic knowledge of the model in an unsupervised manner, but also distinguishes ID/OOD samples to enhance ID classification and OOD detection tasks simultaneously.
\textbf{3)} Extensive experiments demonstrate our approach achieves state-of-the-art performance on SC-OOD benchmarks.
	
\section{Related Works}
	
\noindent\textbf{Out-of-Distribution Detection.}
	OOD detection aims to distinguish test images that come from different distribution comparing to training samples~\cite{baseline}.
	The simplest baseline uses the maximum softmax probability (MSP) to identify OOD samples, which is based on the observation that DNNs tend to produce lower prediction probability for misclassified and OOD inputs~\cite{baseline}.
	Follow-up work uses various techniques to improve MSP. ODIN~\cite{odin} applies temperature scaling on the softmax layer to increase the separation between ID and OOD probabilities. Small perturbations are also introduced into the input space for further improvement.
	Some probabilistic methods attempt to model the distribution of the training samples and use the likelihood, or density, to identify OOD samples~\cite{mahalanobis,S,likelihood}.
	Besides, ensemble methods can also be used to robustify the models~\cite{waic,eloc}.
	Recently, energy scores from the energy-based model are found theoretically consistent with the probability density and suitable for OOD detection~\cite{energyood}.
	All the above methods only rely on ID samples for OOD detection.
	
	Another group of methods for OOD detection utilizes a set of external OOD data,
	based on which the discrepancy between ID and OOD data is learned.
	The baseline work for this branch is OE~\cite{hendrycks18oe}.
	Based on the MSP baseline~\cite{baseline}, a large-scale selected OOD set is introduced as outlier exposure~(OE)
	and an additional objective is included, expecting DNNs to produce uniform softmax scores for extra samples.
	Afterwards, MCD \cite{mcd} proposes a network with two classifiers, which are forced to produce maximum entropy discrepancy for extra OOD samples.
	Some works explore the optimal strategies for external OOD data sampling~\cite{backgroundsample}.
	However, we find two problems with the previous method of using external OOD data:
	\textbf{1)} in the realistic setting, a purified OOD set is difficult to obtain, as ID samples are inevitably introduced and expensive to filter out.
	\textbf{2)} current methods only regard OOD set holistically,
	neglecting the abundant semantic information within the set.
	In this paper, we take a realistic unlabeled set that is a natural ID/OOD mixture, and hope to well explore the knowledge within it.

\noindent\textbf{Deep Clustering.}
Deep clustering is an unsupervised learning method that trains DNNs using the cluster assignments of the resulting features~\cite{deepclustering}.
In this paper, we integrate it into our main proposed unsupervised dual grouping (UDG),
aiming to not only learn visual representations through self-supervised training on both labeled and unlabeled sets,
but also do cluster-wise OOD probability estimation
to filter out ID samples from the unlabeled set described in Section~\ref{S:udg}.

Besides clustering, other unsupervised methods can also be implemented as auxiliary (pretext) tasks, \textit{i.e.}, patch orderings~\cite{patch,jigsaw}, colorization~\cite{colorization}, 
rotation prediction~\cite{rotation} and contrastive learning~\cite{cpc,moco,simclr}.
Although we believe that they can help discover latent knowledge in the unlabeled set and enhance visual representation, their interactive potential on the primary ID filtering task is relatively limited compared with the clustering auxiliary task.

\section{Our Approach}
\label{S:method}
	\begin{figure*}[tp!]
		\begin{center}
			\includegraphics[width=\linewidth]{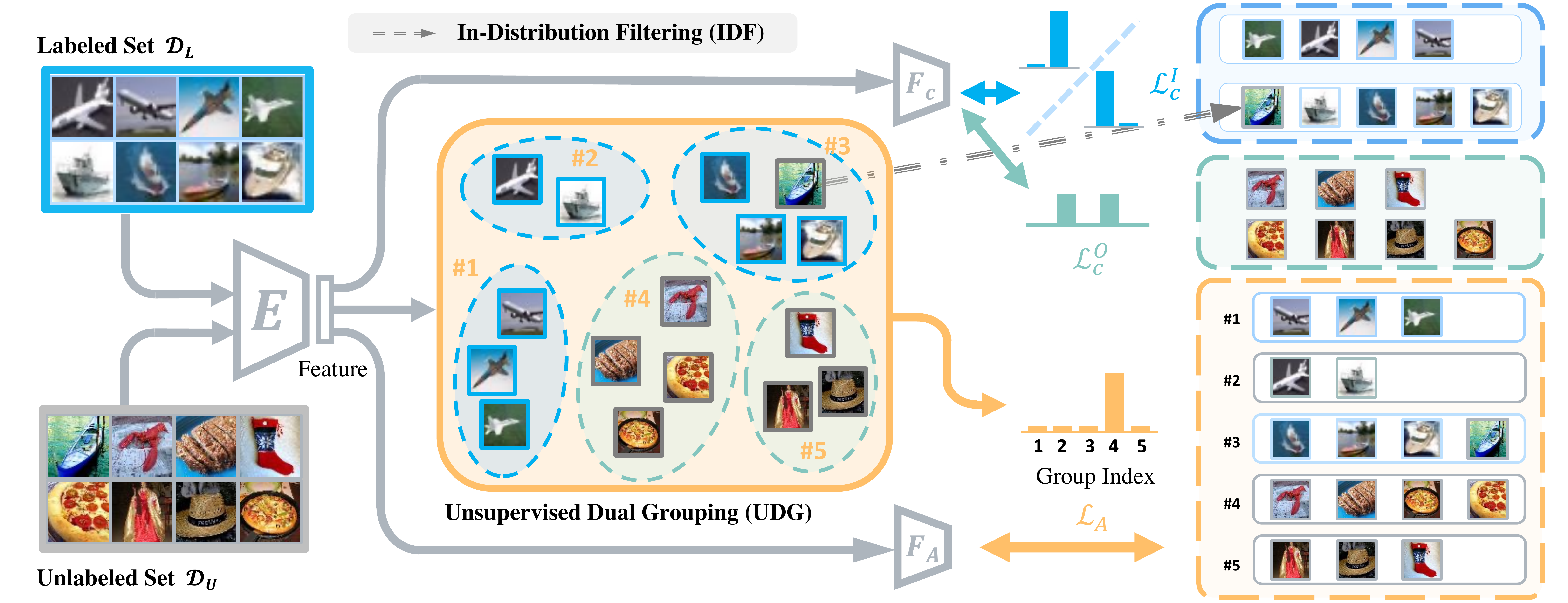}
		\end{center}
		\caption{\textbf{The proposed framework of OOD detection with unsupervised dual grouping~(UDG).} 
			The CNN model has one encoder $E$ and two fully-connected heads $F_C$ and $F_A$. 
			$F_C$ is a classification head, which enables the model to correctly predict ID samples with classification loss $\mathcal{L}_C^I$,
			and to force flatten predictions on unlabeled samples using entropy loss $\mathcal{L}_C^O$.
			$F_A$ is an auxiliary head for deep clustering.
			During the UDG process, the operation of in-distribution filtering~(IDF)
			considers unlabeled samples that fall into ID groups as ID samples 
			for later loss calculation. 
			The group index is then used for the auxiliary deep clustering task.}
		\label{fig:main}
	\end{figure*}
	
	In this section, we introduce our proposed end-to-end pipeline with unsupervised dual grouping~(UDG) in detail.
	
	\subsection{Problem Statement}
	Suppose that we have training set $\mathcal{D}$ and testing set $\mathcal{T}$.
	Under closed-world assumption, both training and testing data are from in-distribution~$\mathcal{I}$, 
	\ie, $\mathcal{D}=\mathcal{D}_L\subset\mathcal{I}$ and $\mathcal{T}=\mathcal{T}^I\subset\mathcal{I}$.
	The subscript $L$ means that $\mathcal{D}_L$ is fully labeled.
	Samples in $\mathcal{T}$~($\mathcal{T}^I$) only belong to known classes $\mathcal{C}^I$ that provided by labels in $\mathcal{D}$~($\mathcal{D}_L$).
	However, a more realistic setting suggests that $\mathcal{T}$ also contains unknown classes $\mathcal{C}^O$ 
	that from out-of-distribution $\mathcal{O}$,
	\ie, $\mathcal{T}$ is composed by $\mathcal{T}^I \subset \mathcal{I}$ and $\mathcal{T}^O \subset \mathcal{O}$.
	A model trained by $\mathcal{D}$ is required to 
	not only correctly classify samples from $\mathcal{T}^I$ into $\mathcal{C}^I$,
	but also recognize OOD samples from $\mathcal{T}^O$.
	To this end, an unlabeled set $\mathcal{D}_U$ is introduced to assist the training process, leading to $\mathcal{D}=\mathcal{D}_L\cup\mathcal{D}_U$.
	Ideally, unlabeled set should be purely from out-of-distribution, \textit{i.e.}~$\mathcal{D}_U\subset \mathcal{O}$. 
	However, in the real practice, 
	$\mathcal{D}_U$ is a mixture of both $\mathcal{D}_U^I \subset \mathcal{I}$ and $\mathcal{D}_U^O \subset \mathcal{O}$
	with unknown separation.
	It is also mentioned that $\mathcal{D}_U^O$ does not necessarily cover $\mathcal{T}^O$.
	In summary, our goal is to train an image classifier from training set $\mathcal{D}=\mathcal{D}_L\cup\mathcal{D}_U$,
	so that the model has the capacity to reject $\mathcal{T}^O$, in addition to classify samples from $\mathcal{T}^I$ correctly.
	\subsection{Framework Overview}
	To empower the classifier with OOD detection ability using both ID set~$\mathcal{D}_L$ and unlabeled set~$\mathcal{D}_U$, 
	our pipeline design is initiated by a classic OE architecture~\cite{hendrycks18oe} 
	that trains the network to classify ID samples correctly 
	and forces high-entropy predictions on unlabeled samples, 
	which is encapsulated in the classifier branch to be introduced in Section~\ref{S:main_task}.
	Then, an unsupervised dual grouping~(UDG) is proposed to group $\mathcal{D}_L$ and $\mathcal{D}_U$ altogether.
	Based on the grouping, ID samples from $\mathcal{D}_U$ can be filtered out 
	and redirected to $\mathcal{D}_L$
	to enhance the performance of classification branch,
	which is introduced in Section~\ref{S:udg}.
	With the grouping information produced by UDG,
	an auxiliary deep clustering branch is attached,
	aiming to discover the valuable but understudied knowledge in the unlabeled set.
	Finally, the entire training and testing procedure is summarized in Section~\ref{S:process}.
	Figure~\ref{fig:main} illustrates the proposed pipeline.
	
	\subsection{Main Task: Classification and Entropy Loss}
	\label{S:main_task}
	We firstly focus on the ID classification ability of the proposed model.
	A classifier is built, containing a backbone encoder~$E$ with learnable parameter $\theta_E$ 
	and a classification head $F_C$ with learnable parameter $\theta_C$.
	A standard cross-entropy loss is utilized to train the classifier using data-label pairs in $\mathcal{D}_L$, 
	formulated by Equation~\ref{E:classification_loss}.
	\begin{equation}
	\label{E:classification_loss}
	\mathcal{L}_C^L = - \frac{1}{\lvert \mathcal{D}_L \rvert} 
	\sum_{(x_i,y_i)\in\mathcal{D}_L} \log \Big( p_{y_i}(y|x_i, \theta_E, \theta_C) \Big)
	\end{equation}
	
	Now, we use the unlabeled set to help the network gain OOD detection capabilities, following the classic architecture in outlier exposure~\cite{hendrycks18oe}.
	Ideally, all samples from the unlabeled set are OOD, \textit{i.e.} $\mathcal{D}_U\subset \mathcal{O}$.
	In this case, since they do not belong to any one of the known classes, 
	the network is forced to produce a uniform posterior distribution over all known classes for unlabeled samples.
	Therefore, an entropy loss is introduced in Equation~\ref{E:entropy_loss} to flatten the model prediction on unlabeled samples~\cite{hendrycks18oe}.
	
	\begin{equation}
	\label{E:entropy_loss}
	\mathcal{L}_C^U = - \frac{1}{\lvert \mathcal{D}_U \rvert} \frac{1}{\lvert \mathcal{C}_I \rvert} 
	\sum_{x_i\in\mathcal{D}_U} \sum_{c\in\mathcal{C}_I} \log \Big( p_c(y|x_i, \theta_E, \theta_C) \Big)
	\end{equation}
	
	However, in real practice, $\mathcal{D}_U$ might be mixed with ID samples, leading to the assumption of $\mathcal{D}_U\subset \mathcal{O}$ inaccurate.
	The problem awaits solving by the proposed unsupervised dual grouping in Section~\ref{S:udg}.
	
	\subsection{Unsupervised Dual Grouping~(UDG)}
	\label{S:udg}
	In this section, we first introduce the basic operation of UDG 
	and then focus on how UDG solves the mentioned problem on entropy loss $\mathcal{L}_C^U$
	when the ID samples are mixed in $\mathcal{D}_U$.
	Ideally, samples in $\mathcal{D}^I_U$ should be removed from $\mathcal{D}_U$
	and return to $\mathcal{D}_L$ with their corresponding labels,
	only leaving $\mathcal{D}^O_U$ for entropy loss minimization.
	Fortunately, UDG has the ability to complete the task 
	with In-Distribution Filtering (IDF) operation.
	Besides, the grouping information provided by UDG 
	can be used by an auxiliary branch of deep clustering to explore the knowledge in the unlabeled data and in turn boost the semantic capability of the model.
	
	\noindent\textbf{Basic Operation.}
	We divide the entire training set $\mathcal{D}$ into $K$ groups for every epoch.
	At the $t$-th epoch, the encoder~$E$ (denoted as~$E^{(t)}$) extracts the feature of every training sample to form a feature set~$\mathcal{F}^{(t)}$ for the following grouping process.
	Any clustering methods can be implemented for grouping, while in this work,
	we use the classic $k$-means algorithm~\cite{kmeans} as in Equation~\ref{E:kmeans},
	where $\mathcal{G}^{(t)}$ restores group index of every sample and
	$g^{(t)}_i\in\mathcal{G}^{(t)}$ denotes the group index of sample $x_i$ at epoch $t$.
	\begin{equation}
	\label{E:kmeans}
	\mathcal{G}^{(t)} \leftarrow kmeans(\mathcal{F}^{(t)}), ~ \text{where}~
	\mathcal{F}^{(t)}=\{E^{(t)}(x)\lvert x \in \mathcal{D} \}
	\end{equation}
	
	Since then, every sample should belong to one of the $K$ groups.
	Formally, all samples that belong to $k$-th group at the $t$-th epoch 
	form the set $\mathcal{D}_k$ in Equation~\ref{E:group}.
	\begin{equation}
    	\label{E:group}
    	\mathcal{D}_{k}^{(t)}=\{x_i\vert g^{(t)}_i=k, (x_i, g^{(t)}_i) \in (\mathcal{D}, \mathcal{G}^{(t)}) \}
	\end{equation}

	\noindent\textbf{In-Distribution Filtering (IDF) with UDG.}
	In this part, we introduce how we filter out ID samples on the scope of groups.
	Empirically, for a group that is dominated by a labeled class,
	the unlabeled data in the group is more likely to belong to the corresponding category.
	With the rule that the ID property of the sample can be estimated
	based on the group they belong to, 
	the operation of In-Distribution Filtering (IDF) is proposed
	to filter out ID samples from the unlabeled set.
	For group $k$, we define its group purity $\gamma^{(t)}_{k,c}$
	to show the proportion of samples belonging to class $c$ at epoch $t$ as Equation~\ref{E:group_purity},
	where $[\mathcal{D}_{L}]_c$ denotes all labeled samples within class $c$.
	\begin{equation}
	\label{E:group_purity}
	\gamma_{k,c}^{(t)}=\frac{\lvert \mathcal{D}_{k}^{(t)} \cap [\mathcal{D}_{L}]_c\rvert}{\lvert \mathcal{D}_{k}^{(t)}\rvert}
	\end{equation}
	
	With group purity $\gamma^{(t)}_{k,c}$ prepared, 
	IDF operator returns all unlabeled samples in a group 
	with group purity over a threshold $\tau$ back to labeled set
	with their labels identical to the group majority,
	forming an updated labeled set $\mathcal{D}_L^{(t)}$
	according to Equation~\ref{E:new_ind}.
	
	\begin{equation}
    \label{E:new_ind}
    	\mathcal{D}_{L}^{(t)}= \mathcal{D}_{L} \cup \{x \vert x \in \mathcal{D}_k^{(t)}, \gamma^{(t)}_{k,c} > \tau\}
	\end{equation}
	Notice that from the second epoch, $\mathcal{D}_{L}$ in Equation~\ref{E:group_purity}
	will be replaced with $\mathcal{D}^{(t-1)}_{L}$, 
	but $\mathcal{D}_{L}$ in Equation~\ref{E:new_ind} remains unchanged 
	to enable an error correction mechanism.
	
	Complementary to the labeled set which is updated into $\mathcal{D}_{L}^{(t)}$, 
	unlabeled set is also updated as $\mathcal{D}_{U}^{(t)}$.
	Using the updated sets, both the classification loss and entropy loss are modified 
	as Equation~\ref{E:new_classification_loss} and Equation~\ref{E:new_entropy_loss}.
	 
	\begin{equation}
	\label{E:new_classification_loss}
	[\mathcal{L}_C^I]^{(t)} = - \frac{1}{\lvert \mathcal{D}^{(t)}_L \rvert} 
	\sum_{(x_i,y_i)\in\mathcal{D}^{(t)}_L} \log \Big( p_{y_i}(y|x_i, \theta_E, \theta_C) \Big)
	\end{equation}
	
	\begin{equation}
	\label{E:new_entropy_loss}
	[\mathcal{L}_C^O]^{(t)} = - \frac{1}{\lvert \mathcal{D}^{(t)}_U \rvert} \frac{1}{\lvert \mathcal{C}_I \rvert} 
	\sum_{x_i\in\mathcal{D}^{(t)}_U} \sum_{c\in\mathcal{C}_I} \log \Big( p_c(y|x_i, \theta_E, \theta_C) \Big)
	\end{equation}
	
	\noindent\textbf{Auxiliary Task for UDG.}
	The motivation for designing the auxiliary branch is to fully leverage the knowledge contained in the unlabeled set, 
	expecting that the learned semantics can further benefit the model performance,
	especially on ID classification. 
	It is expected that the learned semantics can further benefit the model performance, especially on ID classification.
	Fortunately, the groups that UDG provides are perfectly compatible with a deep clustering process~\cite{deepclustering},
	which is used as the unsupervised auxiliary task for knowledge exploration.

	The intuition of deep clustering is that samples that lie in the same group are supposed to be in the same category.
	As the group index for every sample is provided in $\mathcal{G}^{(t)}$ by Equation~\ref{E:kmeans},
	a fully-connected auxiliary head with learnable parameter $\theta_A$ is trained 
	to classify the sample into its corresponding group with auxiliary loss $\mathcal{L}_A$ in Equation~\ref{E:auxiliary_loss}.
	
	\begin{equation}
	\label{E:auxiliary_loss}
	\mathcal{L}^{(t)}_A = - \frac{1}{\lvert \mathcal{D} \rvert} 
	\sum_{(x_i, g_i)\in (\mathcal{D}, \mathcal{G}^{(t)})} \log \Big( p_{g_i}(y|x_i, \theta_E, \theta_A) \Big)
	\end{equation}

	\subsection{Training and Testing Process}
	\label{S:process}
	Finally, with the modified classification loss $[\mathcal{L}_C^I]^{(t)}$, modified entropy loss $[\mathcal{L}_C^O]^{(t)}$ and auxiliary loss $\mathcal{L}^{(t)}_A$,
	the final loss $\mathcal{L}$ can be calculated by Equation~\ref{E:total_loss} with hyperparameter $\lambda_U$ and $\lambda_A$.
	An end-to-end training process is performed to optimize encoder $E$ 
	with parameter $\theta_E$, classification head $F_C$ with parameter $\theta_C$ 
	and auxiliary head $F_A$ with parameter $\theta_A$ simultaneously.
	
	\begin{equation}
	\label{E:total_loss}
	\mathcal{L}^{(t)} = [\mathcal{L}_C^I]^{(t)} + \lambda_U\cdot [\mathcal{L}_C^O]^{(t)} + \lambda_A\cdot \mathcal{L}_A^{(t)}
	\end{equation}
	
	During testing, only classification head $F_C$ along with backbone encoder $E$ is utilized.
	The model will only make an in-distribution prediction if the value of maximum prediction passes a pre-defined threshold $\delta$.
	Otherwise, the sample would be considered as out-of-distribution one. 
	The testing process is formalized by Equation~\ref{E:testing}.
	\begin{equation}
	\label{E:testing}
	pred= 
	\begin{cases}
	\textit{N.A.}, \qquad \text{if } \max p(y|x_i, \theta_E, \theta_C) < \delta,\\
	\arg\max_cp_c(y|x_i, \theta_E, \theta_C), \quad \text{otherwise}.
	\end{cases}
	\end{equation}
\section{SC-OOD Benchmarks}
\label{S:benchmark}

\begin{table}[t]
\caption{\textbf{Tiny-ImageNet classes that are semantically coherent with exemplar CIFAR-10 classes.}
All images in these classes are labeled as ID for SC-OOD benchmarking.}
\label{T:class-filter}
\centering
\begin{tabular}{@{\hskip 3pt}c|l@{\hskip 3pt}}
\toprule
CIFAR-10    & Tiny-ImageNet Classes \\ \midrule
cat & \begin{tabular}[c]{@{}l@{}}
n02802426 \quad Tabby, Tabby Cat\\ 
n02977058 \quad Egyptian Cat\\ 
n04146614 \quad Persian Cat\end{tabular} \\
\midrule
dog	 & \begin{tabular}[c]{@{}l@{}}
n02823428 \quad Golden Retriever\\
n03388043 \quad Chihuahua\\
n02056570 \quad Yorkshire Terrier\\
n03891332 \quad Labrador Tetriever \\
n03042490 \quad German Shepherd\\
n03930313 \quad Standard Poodle\end{tabular} \\
\bottomrule
\end{tabular}
\end{table}

\begin{figure}[t]
\centering
\includegraphics[width=\linewidth]{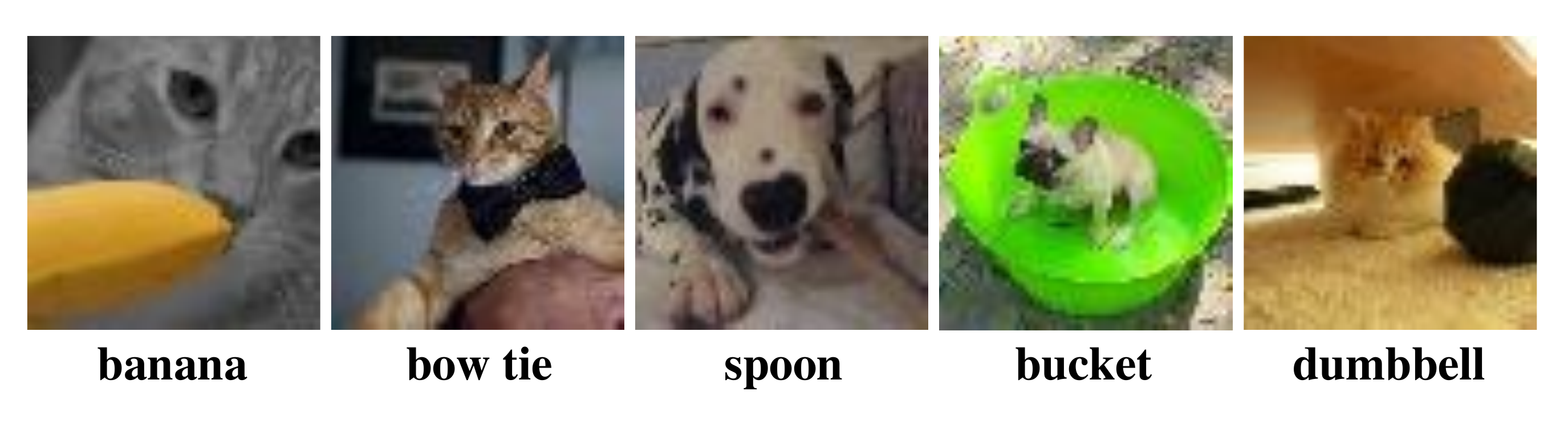}
\caption{\textbf{Exemplar ID images hidden in irrelevant categories are
also filtered for SC-OOD CIFAR-10 benchmark.}
Although the class-wise filtering by Table~\ref{T:class-filter} can identify a large portion of ID samples,
some multi-label images that contains ID semantics requires manually filtering.}
\label{fig:ind-filter}
\end{figure}
In this section, we introduce two benchmarks to reflect semantically coherent OOD detection.
Two benchmarks consider two famous datasets of CIFAR-10/100~\cite{cifar} as in-distribution, respectively. 
Five other datasets including Texture~\cite{texture}, SVHN~\cite{svhn}, Tiny-ImageNet~\cite{tinyImageNet}, LSUN~\cite{lsun}, and Places365~\cite{places365} are prepared as OOD datasets.
We re-split $\mathcal{T}^I$ and $\mathcal{T}^O$ according to the semantics of samples for SC-OOD benchmarks.
There are two steps for re-splitting:
\textbf{1)} we first pick out the ID classes from the OOD datasets and mark all the images within the selected classes as ID samples. Table~\ref{T:class-filter} shows exemplar Tiny-ImageNet ID classes corresponding to two CIFAR-10 classes.
\textbf{2)} we then conduct fine-grained filtering since many images from irrelevant OOD categories also contain ID semantics. Figure~\ref{fig:ind-filter} shows exemplar Tiny-ImageNet ID images that are hidden by irrelevant labels.
Eventually, we obtain CIFAR-10/100 SC-OOD benchmarks with detailed descriptions as follows.

\subsection{Benchmark for CIFAR-10}
CIFAR-10 is a natural object image dataset with 50,000 training samples and 10,000 testing samples from 10 object classes.
Selected datasets for $\mathcal{T}$ include
\textbf{1)}~CIFAR-10 test set with all 10,000 images as $\mathcal{T}^I$;
\textbf{2)}~Entire Texture set with 5,640 images of textural images, all as $\mathcal{T}^O$;
\textbf{3)}~SVHN test set with 26,032 images of real-world street numbers, all as $\mathcal{T}^O$;
\textbf{4)}~CIFAR-100 test set with 10,000 object images that disjoint from CIFAR-10 classes, therefore all as $\mathcal{T}^O$;
\textbf{5)}~Tiny-ImageNet test set containing 10,000 images from 200 objects, with 1,207 images as $\mathcal{T}^I$ and 8,793 images as $\mathcal{T}^O$;
\textbf{6)}~LSUN test set containing 10,000 images for scene recognition,
with 2 images as $\mathcal{T}^I$ and 9,998 images as $\mathcal{T}^O$;
\textbf{7)}~Places365 test set containing 36,500 scene images, with 1,305 images as $\mathcal{T}^I$ and 35,195 images as $\mathcal{T}^O$.

\subsection{Benchmark for CIFAR-100}

CIFAR-100 is a dataset of 100 fine-grained classes with 50,000 training samples and 10,000 testing samples.
The classes between CIFAR-10 and CIFAR-100 are disjoint.
Selected datasets for $\mathcal{T}$ include
\textbf{1)}~CIFAR-100 test set with all 10,000 images as $\mathcal{T}^I$;
\textbf{2)}~Entire Texture set with 5,640 images of textural images, all as $\mathcal{T}^O$;
\textbf{3)}~SVHN test set with 26,032 images of real-world street numbers, all as $\mathcal{T}^O$;
\textbf{4)}~CIFAR-10 test set with 10,000 object images that disjoint from CIFAR-100 classes, therefore all as $\mathcal{T}^O$;
\textbf{5)}~Tiny-ImageNet test set with 2,502 images as $\mathcal{T}^I$ and 7,498 images as $\mathcal{T}^O$;
\textbf{6)}~LSUN test set with 2,429 images as $\mathcal{T}^I$ and 7,571 images as $\mathcal{T}^O$;
\textbf{7)}~Places365 with 2,727 images as $\mathcal{T}^I$ and 33,773 images as $\mathcal{T}^O$.
	
\subsection{Evaluation Metrics}
	We use four kinds of metrics to evaluate the performance on both ID classification and OOD detection.
	
	\textbf{FPR95} is short for the false positive rate (FPR) at 95\% true positive rate (TPR). 
	It measures the portion of falsely recognized OOD when the most of ID samples are recalled.
	
	\textbf{AUROC} computes the area under the receiver operating characteristic curve, evaluating the OOD detection performance. 
	Samples from $\mathcal{T}^I$ are considered positive.
	
	\textbf{AUPR} measures the area under the precision-recall curve.
	Depending on the selection of positiveness, AUPR contains AUPR-In, which regards $\mathcal{T}^I$ as positive, as well as AUPR-Out, where $\mathcal{T}^O$ is considered as positive. In Table~\ref{T:DDOOD} and Table~\ref{T:ablation}, we use AUPR to represent the value of AUPR-Out due to its complementary polarities with AUROC.
	
	\textbf{CCR@FPR$n$} shows Correct Classification Rate~(CCR) at the point when FPR reaches a value $n$.
	The metric evaluates ID classification and OOD detection simultaneously and is formalized by Equation 3 in~\cite{objectosphere}.
	
	Among all the mentioned metrics, only FPR95 is expected to have a lower value on a better model.
	Higher values on any other metrics indicate better performance.

\section{Experiments}
In this section, after describing the implementation details, the effect of each component is analyzed in the ablation study. Then we compare our method with previous state-of-the-art methods. Finally, a more in-depth exploration of our method is discussed.

\noindent\textbf{Experimental Settings.}
Two experimental sets are performed with $\mathcal{D}_L$ of CIFAR-10 and CIFAR-100~\cite{cifar}, respectively.
Both of the training set contains 50,000 images. Tiny-ImageNet~\cite{tinyImageNet} training set is used as $\mathcal{D}_U$ in both experiments. Testing is conducted according to Section~\ref{S:benchmark}.
The main paper only reports the average metric values on all 6 datasets in each benchmark.
All ablation and analytical experiments are performed on the CIFAR-10 benchmark.

\noindent\textbf{Implementation Details.}
All experiments are performed with a standard ResNet-18~\cite{resnet}, trained by an SGD optimizer with a weight decay of $0.0005$ and a momentum of $0.9$. 
Two data-loaders are prepared with batch size of $128$ for $\mathcal{D}_L$ and $256$ for $\mathcal{D}_U$.
A cosine learning rate scheduler is used with an initial learning rate of $0.1$, taking totally $100$ epochs.
For hyperparameters of UDG, we set $\lambda_U=0.5$ and $\lambda_A=0.1$ for all experiments.
Group number $K$ for CIFAR-10/100 is $1000$/$2000$ with IDF threshold $\tau=0.8$.

\begin{figure*}[!t]
\centering
\includegraphics[width=\linewidth]{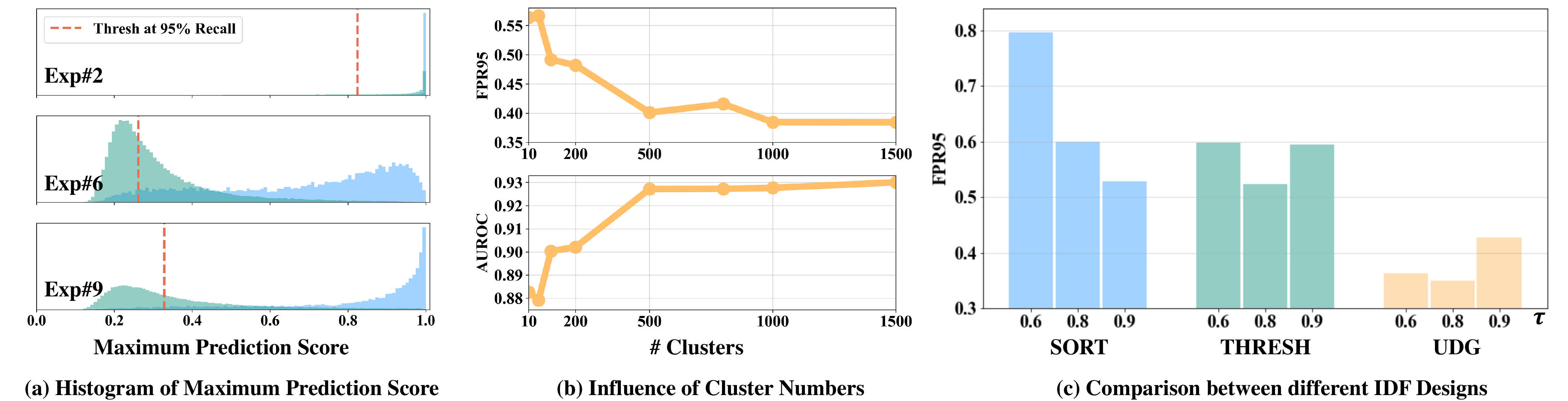}
\caption{\textbf{Comparison and analysis to demonstrate the effectiveness of each module in our framework.}
(a) shows the discrepancy between the maximum prediction scores of ID (blue) and OOD (green) samples in three experiments of Table~\ref{T:ablation} statistically.
(b) shows the the proposed method can obtain a stably good result with a large number of pre-defined clusters.
(c) shows the in-distribution filtering (IDF) strategy we use, \ie UDG, is significantly better than the alternatives.}
\label{fig:exp}
\end{figure*}

\subsection{Ablation Study}

In this section, we will analyze the effect of every major component, including classification task~$\mathcal{L^I_C}$, OE loss~$\mathcal{L^U_C}$, auxiliary deep clustering task~$\mathcal{L_A}$, and in-distribution filtering (IDF) operator $F$ by Table~\ref{T:ablation}.
In summary, OE loss~$\mathcal{L^U_C}$ and IDF operator are shown most effective for OOD detection, 
while the auxiliary deep clustering task can further improve the performance.
Notably, we also report the basic classification accuracy on CIFAR-10 test set, denoted as ACC.
Detailed explanation is conveyed as follows.

\noindent\textbf{Effectiveness of Unlabeled Data.}
Table~\ref{T:ablation} is divided into two major blocks 
according to the usage of unlabeled Tiny-ImageNet.
In this part, we discuss the differences brought by the introduction of unlabeled data.
Exp$\#2$ operates the standard classification where only $\mathcal{L}_C^I$ is used,
and Exp$\#6$ is the standard OE method~\cite{hendrycks18oe} using an additional $\mathcal{L}_U^O$.
The result shows that the OOD detection ability is enhanced by a large margin, as FPR95 gets a significant 7.74\% improvement.
Figure~\ref{fig:exp}\textcolor{red}{-a} compares the histogram of maximum prediction scores between Exp$\#2$ and $\#6$.
The overconfident property is largely reduced in favor of unlabeled samples,
and the ID/OOD discrepancy is also enlarged.
However, it is also worth noting that the ID classification accuracy is reduced 
from 94.94\% to 91.87\%. Since we use a realistic unlabeled set mixing both ID and OOD for OE loss, the unlabeled ID data will wrongly contribute to $\mathcal{L^O_C}$ and therefore harm the classification performance.

\noindent\textbf{Analysis of Unsupervised Dual Grouping.} 
The contribution of UDG is twofold: \textbf{1)}~enabling the IDF operation; \textbf{2)} creating an auxiliary loss $\mathcal{L}_A$.
In this part, we especially focus on the analysis of $\mathcal{L}_A$.
Exp$\#1$ operates a standard deep clustering on CIFAR-10~($K=50$).
The fully-connected layer is finetuned on training set at the end of the training.
A distressing result is obtained on all metrics.
Even worse is Exp$\#4$, which shows that expanding the training set with unlabeled data dominated by OOD further destroys the OOD detection ability of fully unsupervised methods.
Possible explanation is that without the knowing the ID/OOD, deep clustering may easily group ID/OOD samples into one cluster, thus giving up the OOD detection capability.
Therefore, although Exp$\#2$ and $\#3$ show that on the standard CIFAR training set, the auxiliary $\mathcal{L}_A$ can benefit all metrics, the comparison between Exp$\#3$ and Exp$\#5$ unfortunately shows that once an OOD-mixed unlabeled data involved, a simple combination of classification task and unsupervised deep clustering task will damage the OOD detection ability.
Fortunately, after introducing the OOD discrepancy loss $\mathcal{L}_U^O$, the comparison between Exp$\#6$ and Exp$\#7$ shows that the disadvantage of $\mathcal{L}_A$ can be largely reduced, rekindling our hope of regaining the value of $\mathcal{L}_A$.

\begin{table}[t]
	\caption{\textbf{Ablation study on SC-OOD CIFAR-10 benchmark} to show the effectiveness of every component in the proposed framework.
	$F$ represents for the IDF operator.
	For simplicity, we refer to each experiment based on its index (\eg Exp$\#2$ for the experiment with $\mathcal{L_C^I}$ only).}
	\label{T:ablation}
	\centering
	\begin{tabular}{c@{\hskip 5pt}l|@{\hskip 1pt}c@{\hskip 1pt}c@{\hskip 1pt}c@{\hskip 1pt}@{\hskip 1pt}c}
		\toprule
		$\mathcal{D}$ & Components & FPR95 & AUROC & AUPR & ACC \\ 
		\midrule
		\multirow{3}{*}{\rotatebox{90}{CIFAR}}        
		& 1: $\mathcal{L_A}$                   & 89.53 & 65.80 & 65.46 & 64.88 \\
		& 2: $\mathcal{L^I_C}$                 & 58.27 & 89.25 & 87.72 & 94.94 \\
		& 3: $\mathcal{L^I_C} + \mathcal{L_A}$ & \textbf{55.62} & \textbf{90.72} & \textbf{88.33} & \textbf{95.02} \\
		\midrule
		\multirow{6}{*}{\rotatebox{90}{CIFAR+TIN}}        
		& 4: $\mathcal{L_A}$                                       & 91.15 & 64.00 & 63.47 & 59.02  \\
		& 5: $\mathcal{L^I_C} + \mathcal{L_A}$                     & \textbf{62.75} & \textbf{88.21} & \textbf{86.45} & \textbf{94.68}  \\
		\cmidrule{2-6}
		& 6: $\mathcal{L^I_C} + \mathcal{L^U_C}$                   & 50.53 & 88.93 & 87.83 & 91.87  \\
		& 7: $\mathcal{L^I_C} + \mathcal{L^U_C} + \mathcal{L_A}$   & 51.41 & 90.53 & 88.17 & 90.70  \\
		& 8: $\mathcal{L^I_C} + \mathcal{L^U_C} + F$               & 40.93 & 92.23 & 91.92 & 92.34  \\
		& 9: $\mathcal{L^I_C} + \mathcal{L^U_C} + F + \mathcal{L_A}$ & \textbf{36.22} & \textbf{93.78} & \textbf{92.61} & \textbf{92.94}  \\
		\bottomrule
	\end{tabular}
\end{table}

\begin{table*}[btp!]
\setlength{\tabcolsep}{6pt}
\caption{\textbf{Comparison between previous state-of-the-art methods and ours on the SC-OOD CIFAR-10/100 benchmarks.} 
All experiments use ResNet-18~\cite{resnet} for fair comparison. 
ODIN~\cite{odin} and EBO~\cite{energyood} do not require external data, 
and OE~\cite{hendrycks18oe}, MCD~\cite{mcd}, and our UDG use Tiny-ImageNet as unlabeled data.
UDG obtains consistently better results on almost all metrics.}
\label{T:sota}
\centering
\begin{tabular}{cl|ccc|cccc}
\toprule
\multirow{2}{*}{$\mathcal{D}_I$~($\mathcal{D}_U$)} & 
\multirow{2}{*}{Method} 
& \multirow{2}{*}{FPR95~$\downarrow$} 
& \multirow{2}{*}{AUROC~$\uparrow$} 
& \multirow{2}{*}{AUPR(In/Out)~$\uparrow$}
& \multicolumn{4}{c}{CCR@FPR~$\uparrow$} \\ 
\cmidrule(lr){6-9}
& && && $10^{-4}$  & $10^{-3}$  & $10^{-2}$ & $10^{-1}$                      \\ \midrule
\multirow{5}{*}{\begin{tabular}[c]{@{}c@{}}CIFAR-10\\ (Tiny-ImageNet)\end{tabular}}
& ODIN~\cite{odin}         &  52.00 & 82.00 & 73.13~/~85.12 & 0.36 & 1.29& 	6.92    & 39.37 \\ 
& EBO~\cite{energyood}      &  50.03 &	83.83 &	77.15~/~85.11 & 0.49	& 1.93	&9.12	&46.48 \\
& OE~\cite{hendrycks18oe}  & 50.53	& 88.93	& 87.55~/~87.83	& 13.41	& 20.25	& 33.91	& 68.20 \\
& MCD~\cite{mcd}           &73.02&	83.89&	83.39~/~80.53&	5.41&	12.3&	28.02&	62.02  \\
\cmidrule{2-9}
& \textbf{UDG (ours)}   & \textbf{36.22} & \textbf{93.78} & \textbf{93.61}~/~\textbf{92.61} & \textbf{13.87}  & \textbf{34.48} & \textbf{59.97} & \textbf{82.14} \\ 
\midrule
\multirow{5}{*}{\begin{tabular}[c]{@{}c@{}}CIFAR-100\\ (Tiny-ImageNet)\end{tabular}}
& ODIN~\cite{odin}        &81.89	&77.98	&78.54~/~72.56	&1.84	&5.65	& 17.77	& 46.73 \\
& EBO~\cite{energyood}    &81.66	&79.31	&80.54~/~72.82	&2.43	&7.26	& 21.41	& \textbf{49.39} \\
& OE~\cite{hendrycks18oe} &80.06	&78.46	&80.22~/~71.83	&2.74	&8.37	& \textbf{22.18}	& 46.75	\\
& MCD~\cite{mcd}          & 85.14	&74.82	&75.93~/~69.14  &1.06   & 4.60  & 16.73 & 41.83 \\
\cmidrule{2-9}
& \textbf{UDG (ours)}     & \textbf{75.45} & \textbf{79.63}  & \textbf{80.69}~/~\textbf{74.10}  &  \textbf{3.85}  & \textbf{8.66} &  20.57 &  44.47           \\ 
\bottomrule
\end{tabular}
\end{table*}

\noindent\textbf{Effectiveness of In-Distribution Filtering.} 
The contribution of IDF is in twofold: \textbf{1)}~improving ID classification by collecting ID samples from the unlabeled set for better $\mathcal{L}_C^I$, and \textbf{2)} purify from unlabeled ID/OOD mixture into a clean OOD set for better $\mathcal{L}_U^O$.
The comparison between Exp$\#6$ and $\#8$ (UDG but $\lambda_A=0$) illustrates that IDF completes the above goals with a significant 10.4\% benefit on FPR95 and an improvement on classification accuracy. By using a cleaner ID and OOD sets, the auxiliary $\mathcal{L}_A$ finally becomes beneficial in the completed version of Exp$\#9$.

\subsection{Benchmarking Results}
Table~\ref{T:sota} compares our proposed approach with previous state-of-the-art OOD detection methods. Here we only report the average metric values on all 6 OOD datasets for each benchmark due to limited space. Full results are shown in Appendix. Results show that our proposed UDG achieves better results on both SC-OOD benchmarks.

\noindent\textbf{ODIN}~\cite{odin} and \noindent\textbf{Energy-based OOD detector (EBO)}~\cite{energyood} are two representative post-processing OOD methods. We report their best results after hyperparameter searching. 
Their performances are usually inferior to OE method.

\noindent\textbf{Outlier Exposure (OE)}~\cite{hendrycks18oe} corresponds to Exp$\#6$ in Section~\ref{T:ablation}.
The results show that using OOD in training with this mechanism can gain an advantage beyond other baselines.

\noindent\textbf{Maximum Classifier Discrepancy (MCD)} enlarges the entropy discrepancy between two branches to detect OODs~\cite{mcd}. However, we find it significantly overfits the training OOD samples while is difficult to generalize to other OOD domains, leading to disappointing results.

\noindent\textbf{UDG} achieves state-of-the-art results on all metrics of OOD detection. In particular, FPR@95 is significantly reduced on both benchmarks. The full table in the appendix shows that UDG can not only have an advantage on revised SC-OOD datasets such as CIFAR-Places365, but also benefits classic OOD detection test dataset pairs such as CIFAR-Texture, which does not have semantic conflicts.

\subsection{Further Analysis}

\paragraph{Influence of Cluster Numbers.} 
Figure~\ref{fig:exp}\textcolor{red}{-b} shows the influence of the pre-defined cluster number $K$. Generally, increasing $K$ helps converge to the optimal result. When $K$ is small, a larger group size will prevent any group from completely belonging to one class, making it difficult for IDF to filter out any ID samples.
Also, large groups will inevitably include both ID and OOD samples, 
leading the deep clustering task to obscure the ID/OOD discrepancy.
Therefore, our proposed method requires a certain large number of clusters.
Fortunately, experiments show that the results are insensitive to the number of clusters 
when it is large~($K\ge500$), reflecting the practicality of our method.

\paragraph{The Design Choice of IDF.} 
Apart from the IDF proposed in Section~\ref{S:udg}, there are two straightforward alternatives for ID sample filtering.
One solution (denoted as ``THRESH'') filters all unlabeled samples whose maximum softmax scores exceed a predefined threshold $\tau$.
Another solution (called ``SORT'') sorts all unlabeled samples based on their maximum softmax scores and selects the top $(1-\tau)\%$ samples as new ID samples.
We also denote our proposed group-based IDF strategy as ``UDG'', 
which takes all unlabeled samples in a group with ID purity over $\tau$ as ID samples.
Figure~\ref{fig:exp}\textcolor{red}{-c} shows the comparison between them.
Generally, our proposed UDG obtains the best performance on FPR95 comparing to other IDF strategies. SORT achieves the worst performance, since it directly includes a certain number of unlabeled images as ID from the beginning of the end-to-end training. 
The wrongly introduced samples will join classification task so that the error could accumulate, preventing the model from adopting the OOD detection ability properly.
Smaller $\tau$ will let SORT include more unlabeled images without the guarantee of the filtering accuracy.
THRESH has better performance due to its better control of unlabeled samples inclusion.
However, it is still not comparable to UDG, which takes ID samples in a more conservative manner by exploiting a grouping mechanism to reduce overconfidence characteristics of the deep models. 
The results indicate that the group-based ID filtering performs more stably than sample-based methods.

\section{Conclusion}
In this paper, we highlight a problem of current OOD benchmarks
which split ID/OOD according to the data source rather than the semantic meaning, and therefore re-design realistic and challenging SC-OOD benchmarks.
An elegant pipeline named UDG is proposed to achieve the state-of-the-art result on SC-OOD benchmarks, with the usage of realistic unlabeled set.
We hope that the more realistic and challenging SC-OOD setting provides new research opportunities for the OOD community and draw researchers' attention to the importance of data review.

\section*{Acknowledgments}
This work was supported by Innovation and Technology Commission of the Hong Kong Special Administrative Region, China (Enterprise Support Scheme under the Innovation and Technology Fund B/E030/18), NTU NAP, and RIE2020 Industry Alignment Fund – Industry Collaboration Projects (IAF-ICP) Funding Initiative, as well as cash and in-kind contribution from the industry partner(s).

\clearpage
{\small
\bibliographystyle{ieeetr}
\bibliography{final}}
\newpage

\appendix
\setcounter{table}{0}
\renewcommand{\thetable}{A\arabic{table}}
\setcounter{figure}{0}
\renewcommand{\thefigure}{A\arabic{figure}}
\section{More Details on SC-OOD Benchmarking}
\begin{figure}[b]
\centering
\includegraphics[width=0.9\linewidth]{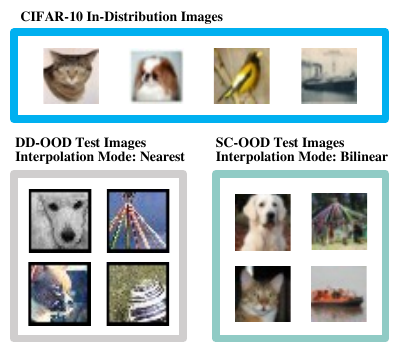}
\caption{\textbf{Exemplar DD-OOD and SC-OOD testing images.} DD-OOD usually utilizes nearest interpolation mode for resizing,
which generates grainy images with some sensory differences compared to ID images. SC-OOD takes bi-linear interpolation mode, yielding a more challenging task to encourage SC-OOD methods to focus on semantics.}
\label{fig:appendix_interpolation}
\end{figure}
\begin{table}[b]
\setlength{\tabcolsep}{3.6pt}
\centering
\caption{\textbf{The record of OOD detection performance as benchmarks gradually changes from DD-OOD to SC-OOD.}
It records totally 4 steps from DD-OOD benchmark of CIFAR-10 + Tiny-ImageNet~(test, nearest interpolation) to SC-OOD using CIFAR-10 + Tiny-ImageNet (val, bi-linear interpolation) after semantics-based re-splitting.}
\label{T:appendix_DDOOD}
\begin{tabular}{l|c|c}
\toprule
 & FPR95~$\downarrow$ & AUROC~$\uparrow$    \\
\midrule
ODIN & 0.46 - 14.3 - 49.9 - 55.0 & 99.8 - 97.3 - 88.3 - 88.8\\
EBO  & 1.56 - 22.8 - 45.6 - 50.6 & 99.5 - 95.9 - 90.2 - 90.4\\
MCD  & 0.01 - 59.1 - 61.5 - 68.6 & 99.9 - 93.3 - 89.3 - 88.9\\
\midrule
UDG  & 12.3 - 18.3 - 43.7 - 48.3 & 97.9 - 96.7 - 91.0 - 91.1\\
\bottomrule
\end{tabular}
\end{table}

In this section, we will explain more details of the SC-OOD benchmark formation mentioned in Section~\ref{S:benchmark}.
We first comprehensively describe the difference between the proposed SC-OOD benchmark and DD-OOD benchmark.
We scrutinized the existing famous OOD detection benchmarks (referred as DD-OOD) and find that they actually utilize nearest interpolation methods when resizing OOD images into ID image size. As shown in Figure~\ref{fig:appendix_interpolation}, DD-OOD images look more coarse and grainy than ID images, resulting in a detectable sensory difference between `smooth' ID images and `coarse' OOD images. In this case, OOD detection methods targeting on DD-OOD benchmark could just impractically focus on low-level covariate shifts and ignore the high-level semantic differences for final decision.
Therefore, we aim to propose a more challenging SC-OOD task to actually focus on semantics. In SC-OOD benchmarks, we use the alternative bi-linear interpolation method for resizing, which yields smoother images that are more similar to ID images. We believe it will encourage the models to focus more on semantics for OOD detection, reflecting the purpose of the SC-OOD benchmark.
Afterward, we redirect the ID samples from OOD datasets, which has been explained in Section~\ref{S:benchmark}.

In sum, two steps from DD-OOD to SC-OOD: \textbf{1)}~using bi-linear interpolation instead of nearest for resizing; \textbf{2)}~re-splitting ID and OOD test sets according to semantics.

Table~\ref{T:appendix_DDOOD} shows the performance changes from DD-OOD to SC-OOD on CIFAR-10 + Tiny-ImageNet~(TIN). Four states are recorded as OOD TIN set gradually changes:
\textbf{1)}~TIN test set, nearest~(interpolation), \textbf{2)}~TIN val set, nearest, \textbf{3)}~TIN val set, bi-linear, \textbf{4)}~TIN val set, bi-linear, with re-splitting as SC-OOD eventually.
We use TIN val set because it contains ground-truth labels for easier re-splitting.
The result shows that even changing test set into validation set will break the perfect performance of some existing methods. Bigger drop exists when interpolation methods change. This drop is understandable since the same interpolation will eliminate all major covariate shifts, but ID and OOD are not yet separated by semantics.
However, semantic re-splitting continues to destroy model performance,
but UDG gets minimal decrease and better overall scores on both metrics, showing a better understanding of semantics.

\begin{figure}[h]
\centering
\includegraphics[width=\linewidth]{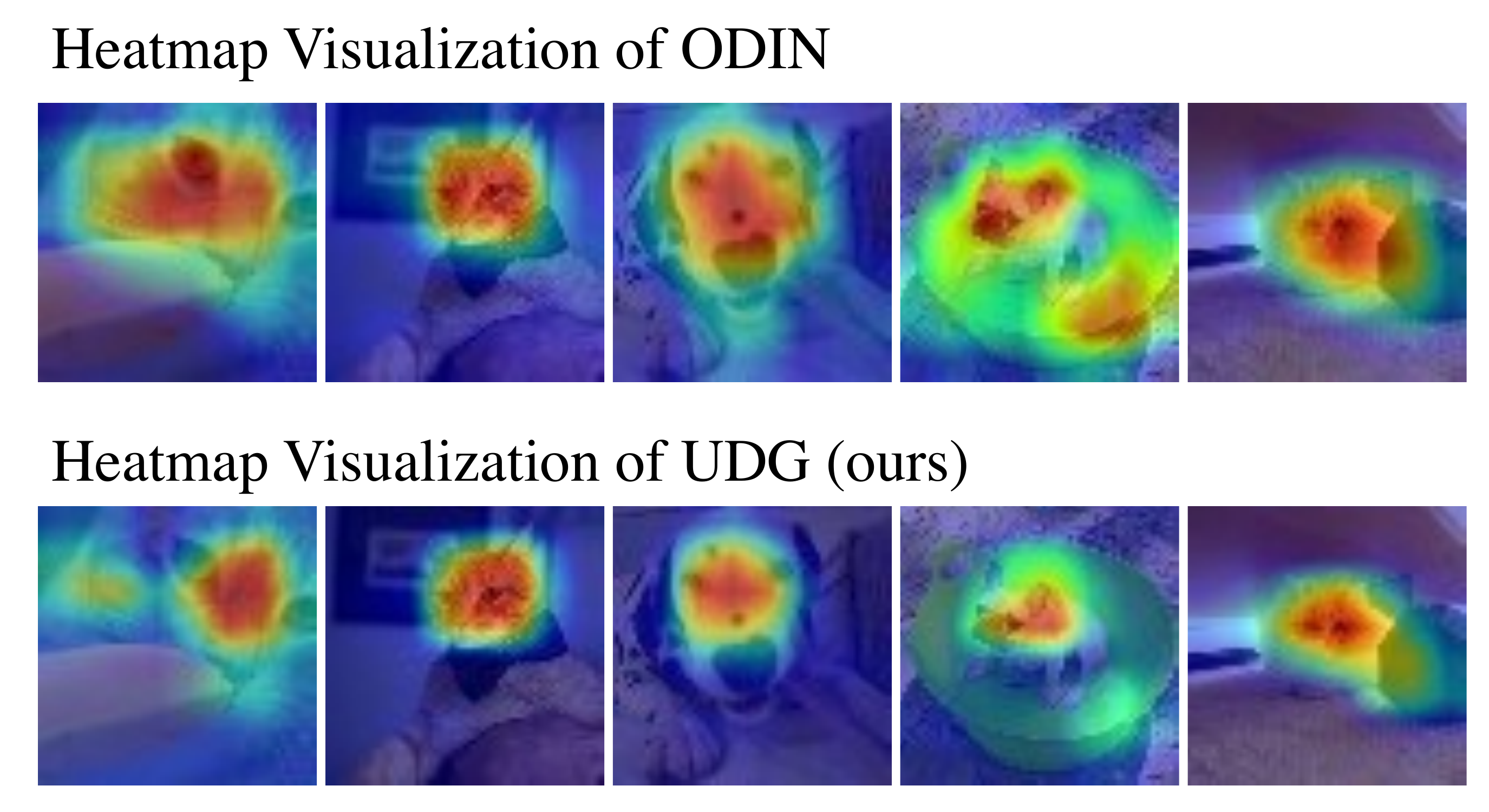}
\caption{\textbf{Heatmap visualization on the images from Figure~\ref{fig:ind-filter}.}
The upper part is from ODIN and the lower part is ours.
For the fourth image of the dog in the bucket, ODIN is distracted by the irrelevant green bucket for its prediction of dog while ours does not distract.
Generally, our method shows better concentration on semantics.}
\label{fig:cam1}
\end{figure}

\section{Visual Heatmap Comparison}
In this section, we visualize the heatmap activated by the 
previous method ODIN~\cite{odin} and our proposed UDG on their prediction.
We found that the semantic capabilities of UDG are significantly stronger than ODIN, since our model can focus more on the semantic area of the image, 
while ODIN usually distracts, sometimes even focuses on some irrelevant area (fourth image in Figure~\ref{fig:cam1}).

\begin{figure}[t]
\centering
\includegraphics[width=\linewidth]{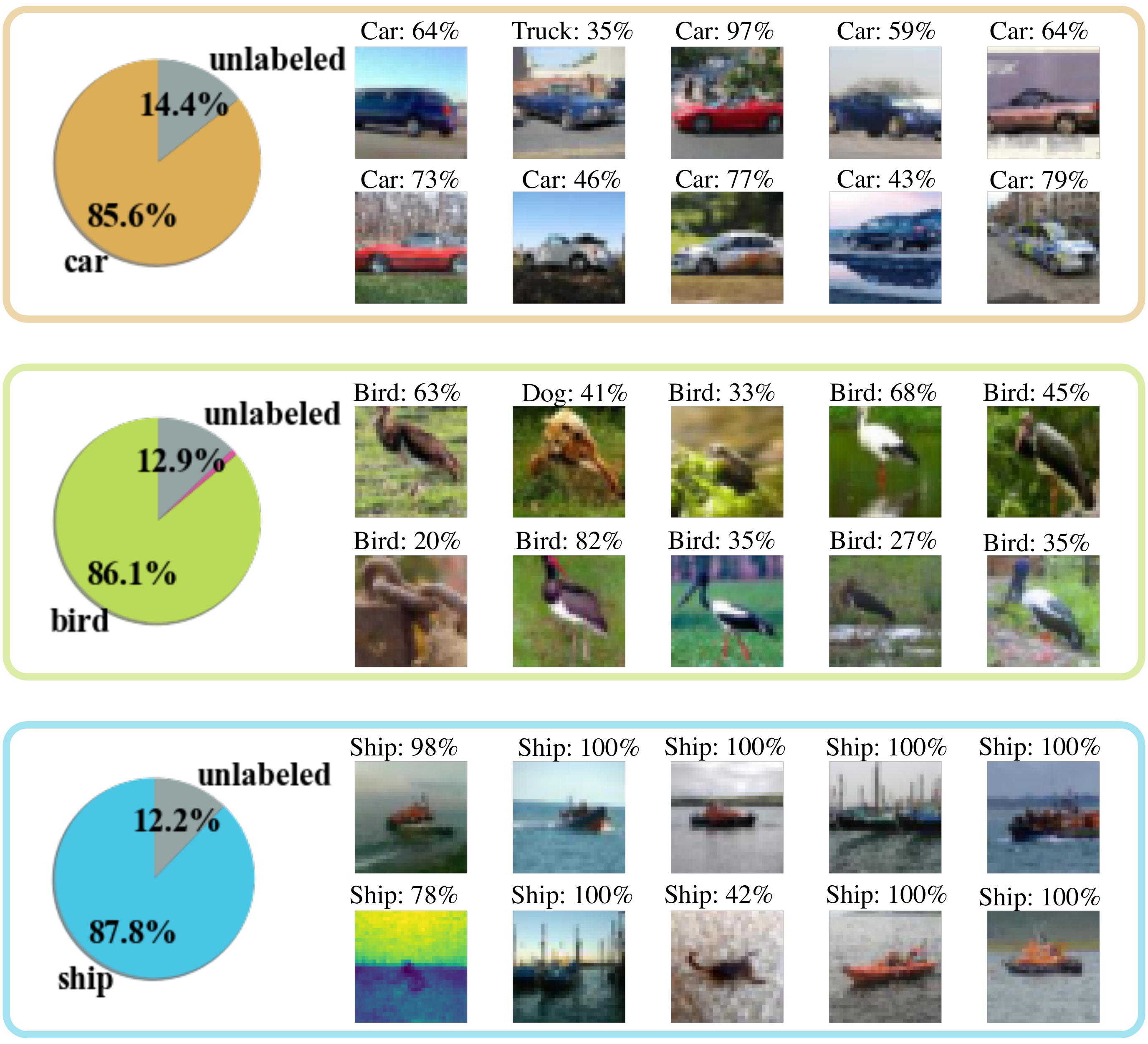}
\caption{\textbf{Visualization of three high group purity clusters for in-distribution filtering (IDF).} 
We randomly show three clusters with group purity over 0.8 at 80\% of the training time.
The visualization shows that our IDF strategy accompanied by UDG
can filter out ID samples in an accurate manner.
The confidence (softmax) score is also presented above each image.
Our group-based IDF strategy can also include ID samples with a lower individual confidence score (refer to the last two images of birds).}
\label{fig:idf_vis}
\end{figure}

\section{Visualization of the Proposed IDF}
In this section, we visualize the in-distribution filtering (IDF) process
that is described in Section~\ref{S:udg}.
According to Figure~\ref{fig:idf_vis}, 
we find that the major unlabeled data that falls in the high-purity ID group
is actually ID samples that belong to the corresponding category.
We also notice that this method can also include these images with a relatively low confidence score,
for example, for the cluster of birds, the last two images only have confidence around 0.3,
which might be difficult to be filtered as ID if we only consider the confidence score.
However, our method would be able to include them.
According to the second image of cars, even though the network provides an incorrect pseudo label of truck for the car,
our IDF strategy can correct the mistake.
Even though there are also few mistakes introduced (scorpion in the ship's group), 
it will be corrected when re-grouping in the next epoch.
In addition, the overconfidence property of neural networks might give a high confidence score 
for wrong images, while the filtering strategy of UDG can also help prevent this mistake.
In sum, the detailed visualization shows the reliability of the proposed IDF method.

\section{Detailed Results and More Architectures}
Table~\ref{T:appendix_cifar10_res18} and Table~\ref{T:appendix_cifar100_res18} expand the average values reported in Table~\ref{T:sota}.
We also do experiments on another network architecture of WideResNet-28~\cite{resnet}.
The result generally has the same trend as ResNet-18 architecture.
The proposed UDG method has advantages on almost all the metrics, showing that our method enhances ID classification and OOD detection ability. Notably, the advantages of our proposed method on Tiny-ImageNet, LSUN, and Places365 largely contribute to the good mean performance of all OOD detection metrics. We consider the above few datasets are difficult samples in the benchmark since many objects have similar but different semantics. A good result is also achieved on easy datasets of Texture and SVHN.

\section{Valuable Comments from Rebuttal}
Here posts an answer we highlighted during the rebuttal period to help readers better understand our paper.

\medskip
\noindent\textbf{[On Motivation of SC-OOD]}
Classic OOD detection aims to train a `conservative' model to distinguish samples 
with either a covariate shift on source distribution~$p(x)$ 
or a semantic shift on label distribution~$p(y)$.
However, we notice an impractical goal of classic OOD detection:
to perfectly distinguish CIFAR cars from ImageNet cars,
even though their covariate shift is negligible.
The unrealistic goal will unfortunately 
result in an extremely narrow range of capabilities for deployed models, 
greatly limiting their use in real applications such as autonomous cars.
In an attempt to address this problem, we form a new, realistic, and challenging SC-OOD task 
that is \textbf{juxtaposed} to classic OOD detection.
SC-OOD re-defines the `distribution' as label distribution~$p(y)$ only
instead of the classic $p(x,y)$.
Under the SC-OOD setting, models are required to:
1) well detect images from different label distributions, 
2) correctly classify images within the same label distribution with negligible source distribution shifts, 
which is consistent with a popular research topic called robustness of deep learning.


\section{Discussion on Drawbacks}
Here we list our current shortcomings.
Although the use of UDG mostly helps alleviate the classification decline of the OE method, it can not yet exceed the standard ID classification performance. More exploration is needed for better use of unlabeled data to achieve stronger ID classification while retaining OOD detection capabilities. Also, we will attempt to analyze UDG on larger datasets such as ImageNet with high-resolution images and complex semantics.

\begin{table*}[btp!]
\setlength{\tabcolsep}{6pt}
\caption{\textbf{Performance details on CIFAR-10 benchmark using ResNet-18.}
UDG obtains consistently better results across OOD detection metrics. Accuracy shows the classification accuracy on all the (filtered) ID test samples, which can be improved by UDG on the top of OE method.}
\label{T:appendix_cifar10_res18}
\centering
\begin{tabular}{c|c|ccc|cccc|c}
\toprule
\multirow{2}{*}{Method} & \multirow{2}{*}{Dataset} 
& \multirow{2}{*}{FPR95~$\downarrow$} 
& \multirow{2}{*}{AUROC~$\uparrow$} 
& \multirow{2}{*}{AUPR(In/Out)~$\uparrow$}
& \multicolumn{4}{c|}{CCR@FPR~$\uparrow$} 
& \multirow{2}{*}{Accuracy~$\uparrow$} \\ 
\cmidrule(lr){6-9}
& && && $10^{-4}$  & $10^{-3}$  & $10^{-2}$ & $10^{-1}$ \\ 
\midrule
\multirow{7}{*}{\begin{tabular}[c]{@{}c@{}}MSP\end{tabular}}

&Texture     & 52.27          & 90.81          & 94.07          ~/~ 82.32          & 0.10          & 1.32          & 20.84          & 79.77          & 95.02          \\
&SVHN     & 50.25          & 92.65          & 87.54          ~/~ 95.84          & 2.47          & 10.73         & 48.22          & 83.96          & 95.02          \\
& CIFAR-100         & 61.19          & 87.40          & 86.30          ~/~ 85.35          & 0.07          & 1.72          & 12.30          & 69.56          & 95.02          \\
& Tiny-ImageNet & 65.32          & 87.32          & 89.41          ~/~ 81.17          & 0.40          & 2.44          & 14.16          & 71.86          & 92.54          \\
& LSUN &  58.62          & 89.34          & 89.30          ~/~ 86.99          & 0.88          & 3.53          & 19.31          & 76.46          & 95.02          \\
& Places365 &  61.99          & 87.96          & 72.61          ~/~ 94.64          & 0.74          & 2.86          & 15.63          & 72.72          & 93.87          \\
\cmidrule{2-10}
& \textbf{Mean} & \textbf{58.27} & \textbf{89.25} & \textbf{86.54} ~/~ \textbf{87.72} & \textbf{0.78} & \textbf{3.77} & \textbf{21.74} & \textbf{75.72} & \textbf{94.42} \\

\midrule
\multirow{7}{*}{\begin{tabular}[c]{@{}c@{}}ODIN\end{tabular}}

&Texture     & 42.52          & 84.06          & 86.01          ~/~ 80.73          & 0.02          & 0.18          & 3.71          & 40.14          & 95.02          \\
&SVHN     & 52.27          & 83.26          & 63.76          ~/~ 92.60          & 1.01          & 4.00          & 11.82         & 44.85          & 95.02          \\
& CIFAR-100         & 56.34          & 78.40          & 73.21          ~/~ 80.99          & 0.10          & 0.38          & 4.43          & 30.11          & 95.02          \\
& Tiny-ImageNet & 59.09          & 79.69          & 79.34          ~/~ 77.52          & 0.36          & 0.63          & 4.49          & 34.52          & 92.54          \\
& LSUN &  47.85          & 84.56          & 81.56          ~/~ 85.58          & 0.21          & 0.85          & 9.92          & 46.95          & 95.02          \\
& Places365 &  53.94          & 82.01          & 54.92          ~/~ 93.30          & 0.47          & 1.68          & 7.13          & 39.63          & 93.87          \\
\cmidrule{2-10}
& \textbf{Mean} & \textbf{52.00} & \textbf{82.00} & \textbf{73.13} ~/~ \textbf{85.12} & \textbf{0.36} & \textbf{1.29} & \textbf{6.92} & \textbf{39.37} & \textbf{94.42} \\

\midrule
\multirow{7}{*}{\begin{tabular}[c]{@{}c@{}}EBO\end{tabular}}

&Texture     & 52.11          & 80.70          & 83.34          ~/~ 75.20          & 0.01          & 0.13          & 2.79          & 31.96          & 95.02          \\
&SVHN     & 30.56          & 92.08          & 80.95          ~/~ 96.28          & 1.85          & 5.74          & 21.44         & 75.81          & 95.02          \\
& CIFAR-100         & 56.98          & 79.65          & 75.09          ~/~ 81.23          & 0.10          & 0.69          & 4.74          & 34.28          & 95.02          \\
& Tiny-ImageNet & 57.81          & 81.65          & 81.80          ~/~ 78.75          & 0.33          & 0.95          & 6.01          & 40.40          & 92.54          \\
& LSUN &  50.56          & 85.04          & 82.80          ~/~ 85.29          & 0.24          & 1.96          & 11.35         & 50.43          & 95.02          \\
& Places365 &  52.16          & 83.86          & 58.96          ~/~ 93.90          & 0.39          & 2.11          & 8.38          & 46.00          & 93.87          \\
\cmidrule{2-10}
& \textbf{Mean} & \textbf{50.03} & \textbf{83.83} & \textbf{77.15} ~/~ \textbf{85.11} & \textbf{0.49} & \textbf{1.93} & \textbf{9.12} & \textbf{46.48} & \textbf{94.42} \\

\midrule
\multirow{7}{*}{\begin{tabular}[c]{@{}c@{}}MCD\end{tabular}}

&Texture     & 83.92          & 81.59          & 90.20           ~/~ 63.27          & 4.97          & 10.51         & 29.52          & 62.10           & 90.56          \\
&SVHN     & 60.27          & 89.78          & 85.33          ~/~ 94.25          & 20.05         & 38.23         & 55.43          & 74.01          & 90.56          \\
& CIFAR-100         & 74.00             & 82.78          & 83.97          ~/~ 79.16          & 0.80           & 4.99          & 18.88          & 58.18          & 90.56          \\
& Tiny-ImageNet & 78.89          & 80.98          & 85.63          ~/~ 72.48          & 1.62          & 4.15          & 19.37          & 56.08          & 87.33          \\
& LSUN &  68.96          & 84.71          & 85.74          ~/~ 81.50           & 1.75          & 7.93          & 21.88          & 61.54          & 90.56          \\
& Places365 &  72.08          & 83.51          & 69.44          ~/~ 92.52          & 3.29          & 7.97          & 23.07          & 60.22          & 88.51          \\
\cmidrule{2-10}
& \textbf{Mean} & \textbf{73.02} & \textbf{83.89} & \textbf{83.39} ~/~ \textbf{80.53} & \textbf{5.41} & \textbf{12.30} & \textbf{28.02} & \textbf{62.02} & \textbf{89.68} \\

\midrule
\multirow{7}{*}{\begin{tabular}[c]{@{}c@{}}OE\end{tabular}}

&Texture     & 51.17          & 89.56          & 93.79          ~/~ 81.88          & 6.58           & 11.80           & 27.99          & 71.13         & 91.87          \\
&SVHN     & 20.88          & 96.43          & 93.62          ~/~ 98.32          & 32.72          & 47.33          & 67.20           & 86.75         & 91.87          \\
& CIFAR-100         & 58.54          & 86.22          & 86.17          ~/~ 84.88          & 3.64           & 6.55           & 19.04          & 61.11         & 91.87          \\
& Tiny-ImageNet & 58.98          & 87.65          & 90.9           ~/~ 82.16          & 14.37          & 18.84          & 33.65          & 66.03         & 89.27          \\
& LSUN &  57.97          & 86.75          & 87.69          ~/~ 85.07          & 11.8           & 19.62          & 29.22          & 61.95         & 91.87          \\
& Places365 &  55.64          & 87.00             & 73.11          ~/~ 94.67          & 11.36          & 17.36          & 26.33          & 62.23         & 90.99          \\
\cmidrule{2-10}
& \textbf{Mean} & \textbf{50.53} & \textbf{88.93} & \textbf{87.55} ~/~ \textbf{87.83} & \textbf{13.41} & \textbf{20.25} & \textbf{33.91} & \textbf{68.20} & \textbf{91.29} \\

\midrule
\multirow{7}{*}{\begin{tabular}[c]{@{}c@{}}UDG\end{tabular}}

&Texture     & 20.43          & 96.44          & 98.12          ~/~ 92.91          & 19.90           & 43.33          & 69.19          & 87.71          & 92.94          \\
&SVHN     & 13.26          & 97.49          & 95.66          ~/~ 98.69          & 36.64          & 56.81          & 76.77          & 89.54          & 92.94          \\
& CIFAR-100         & 47.20           & 90.98          & 91.74          ~/~ 89.36          & 1.50            & 10.94          & 40.34          & 75.89          & 92.94          \\
& Tiny-ImageNet & 50.18          & 91.91          & 94.43          ~/~ 86.99          & 0.32           & 23.15          & 53.96          & 78.36          & 90.22          \\
& LSUN &  42.05          & 93.21          & 94.53          ~/~ 91.03          & 14.26          & 37.59          & 60.62          & 81.69          & 92.94          \\
& Places365 &  44.22          & 92.64          & 87.17          ~/~ 96.66          & 10.62          & 35.05          & 58.96          & 79.63          & 91.68          \\
\cmidrule{2-10}
& \textbf{Mean} & \textbf{36.22} & \textbf{93.78} & \textbf{93.61} ~/~ \textbf{92.61} & \textbf{13.87} & \textbf{34.48} & \textbf{59.97} & \textbf{82.14} & \textbf{92.28} \\

\bottomrule
\end{tabular}
\end{table*}

\begin{table*}[btp!]
\setlength{\tabcolsep}{6pt}
\caption{\textbf{Performance details on CIFAR-100 benchmark using ResNet-18.}
UDG obtains consistently better results across OOD detection metrics. Accuracy shows the classification accuracy on all the (filtered) ID test samples.}
\label{T:appendix_cifar100_res18}
\centering
\begin{tabular}{c|c|ccc|cccc|c}
\toprule
\multirow{2}{*}{Method} & \multirow{2}{*}{Dataset} 
& \multirow{2}{*}{FPR95~$\downarrow$} 
& \multirow{2}{*}{AUROC~$\uparrow$} 
& \multirow{2}{*}{AUPR(In/Out)~$\uparrow$}
& \multicolumn{4}{c|}{CCR@FPR~$\uparrow$} 
& \multirow{2}{*}{Accuracy~$\uparrow$} \\ 
\cmidrule(lr){6-9}
& && && $10^{-4}$  & $10^{-3}$  & $10^{-2}$ & $10^{-1}$ \\ 
\midrule
\multirow{7}{*}{\begin{tabular}[c]{@{}c@{}}MSP\end{tabular}}

&Texture     & 84.04          & 75.85          & 85.72          ~/~ 58.63          & 0.41          & 3.67          & 16.26          & 45.84          & 76.65          \\
&SVHN     & 80.12          & 80.01          & 70.84          ~/~ 88.52          & 9.90          & 17.77         & 31.00          & 52.94          & 76.65          \\
& CIFAR-10    & 80.64          & 78.33          & 80.69          ~/~ 74.04          & 0.00          & 5.94          & 21.09          & 49.10          & 76.65          \\
& Tiny-ImageNet & 83.32          & 77.85          & 86.97          ~/~ 61.73          & 2.43          & 7.55          & 24.69          & 48.29          & 69.56          \\
& LSUN &  83.03          & 77.31          & 86.31          ~/~ 1.45          & 3.38          & 6.73          & 21.49          & 47.88          & 76.10          \\
& Places365 &  77.57          & 79.99          & 67.55          ~/~ 89.21          & 1.11          & 6.02          & 22.72          & 51.69          & 77.56          \\
\cmidrule{2-10}
& \textbf{Mean} & \textbf{81.45} & \textbf{78.22} & \textbf{79.68} ~/~ \textbf{72.26} & \textbf{2.87} & \textbf{7.95} & \textbf{22.88} & \textbf{49.29} & \textbf{75.53} \\
\midrule
\multirow{7}{*}{\begin{tabular}[c]{@{}c@{}}ODIN\end{tabular}}

&Texture     & 79.47          & 77.92          & 86.69          ~/~ 62.97          & 2.66          & 4.66          & 15.09          & 45.82          & 76.65          \\
&SVHN     & 90.33          & 75.59          & 65.25          ~/~ 84.49          & 4.98          & 12.02         & 23.79          & 46.61          & 76.65          \\
& CIFAR-10    & 81.82          & 77.90          & 79.93          ~/~ 73.39          & 0.09          & 3.69          & 15.39          & 47.20          & 76.65          \\
& Tiny-ImageNet & 82.74          & 77.58          & 86.26          ~/~ 61.38          & 0.20          & 3.78          & 15.99          & 45.56          & 69.56          \\
& LSUN &  80.57          & 78.22          & 86.34          ~/~ 63.44          & 1.68          & 5.59          & 17.37          & 45.56          & 76.10          \\
& Places365 &  76.42          & 80.66          & 66.77          ~/~ 89.66          & 1.45          & 4.16          & 18.98          & 49.60          & 77.56          \\
\cmidrule{2-10}
& \textbf{Mean} & \textbf{81.89} & \textbf{77.98} & \textbf{78.54} ~/~ \textbf{72.56} & \textbf{1.84} & \textbf{5.65} & \textbf{17.77} & \textbf{46.73} & \textbf{75.53} \\
\midrule
\multirow{7}{*}{\begin{tabular}[c]{@{}c@{}}EBO\end{tabular}}

& Texture     & 84.29          & 76.32          & 85.87          ~/~ 59.12          & 0.82          & 3.89          & 14.37          & 44.60          & 76.65          \\
&SVHN     & 78.23          & 83.57          & 75.61          ~/~ 90.24          & 9.67          & 17.27         & 33.70          & 57.26          & 76.65          \\
& CIFAR-10    & 81.25          & 78.95          & 80.01          ~/~ 74.44          & 0.05          & 4.63          & 18.03          & 48.67          & 76.65          \\
& Tiny-ImageNet & 83.32          & 78.34          & 87.08          ~/~ 62.13          & 1.04          & 6.37          & 21.44          & 47.92          & 69.56          \\
& LSUN &  84.51          & 77.66          & 86.42          ~/~ 61.40          & 1.59          & 6.44          & 19.58          & 46.66          & 76.10          \\
& Places365 &  78.37          & 80.99          & 68.22          ~/~ 89.60          & 1.40          & 4.94          & 21.32          & 51.21          & 77.56          \\
\cmidrule{2-10}
& \textbf{Mean} & \textbf{81.66} & \textbf{79.31} & \textbf{80.54} ~/~ \textbf{72.82} & \textbf{2.43} & \textbf{7.26} & \textbf{21.41} & \textbf{49.39} & \textbf{75.53} \\
\midrule  
\multirow{7}{*}{\begin{tabular}[c]{@{}c@{}}MCD\end{tabular}}
&Texture     & 83.97          & 73.46          & 83.11          ~/~ 56.79          & 0.07          & 1.03          & 9.29           & 38.09          & 68.80          \\
&SVHN     & 85.82          & 76.61          & 65.50          ~/~ 85.52          & 3.03          & 8.66          & 23.15          & 45.44          & 68.80          \\
& CIFAR-10    & 87.74          & 73.15          & 76.51          ~/~ 67.24          & 0.35          & 3.26          & 16.18          & 41.41          & 68.80          \\
& Tiny-ImageNet  & 84.46          & 75.32          & 85.11          ~/~ 59.49          & 0.24          & 6.14          & 19.66          & 41.44          & 62.21          \\
& LSUN & 86.08          & 74.05          & 84.21          ~/~ 58.62          & 1.57          & 5.16          & 18.05          & 41.25          & 67.51          \\
& Places365 & 82.74          & 76.30          & 61.15          ~/~ 87.19          & 1.08          & 3.35          & 14.04          & 43.37          & 70.47          \\
\cmidrule{2-10}
& \textbf{Mean} & \textbf{85.14} & \textbf{74.82} & \textbf{75.93} ~/~ \textbf{69.14} & \textbf{1.06} & \textbf{4.60} & \textbf{16.73} & \textbf{41.83} & \textbf{67.77} \\
\midrule
\multirow{7}{*}{\begin{tabular}[c]{@{}c@{}}OE\end{tabular}}
&Texture     & 86.56          & 73.89          & 84.48          ~/~ 54.84          & 0.66          & 2.86          & 12.86          & 41.81          & 70.49          \\
&SVHN     & 68.87          & 84.23          & 75.11          ~/~ 91.41          & 7.33          & 14.07         & 31.53          & 54.62          & 70.49          \\
& CIFAR-10    & 79.72          & 78.92          & 81.95          ~/~ 74.28          & 2.82          & 9.53          & 23.90          & 48.21          & 70.49          \\
& Tiny-ImageNet & 83.41          & 76.99          & 86.36          ~/~ 60.56          & 0.22          & 8.50          & 21.95          & 43.98          & 63.69          \\
& LSUN &  83.53          & 77.10          & 86.28          ~/~ 60.97          & 1.72          & 7.91          & 22.61          & 44.19          & 69.89          \\
& Places365 &  78.24          & 79.62          & 67.13          ~/~ 88.89          & 3.69          & 7.35          & 20.22          & 47.68          & 72.02          \\
\cmidrule{2-10}
& \textbf{Mean} & \textbf{80.06} & \textbf{78.46} & \textbf{80.22} ~/~ \textbf{71.83} & \textbf{2.74} & \textbf{8.37} & \textbf{22.18} & \textbf{46.75} & \textbf{69.51} \\
\midrule
\multirow{7}{*}{\begin{tabular}[c]{@{}c@{}}UDG\end{tabular}}
&Texture     & 75.04          & 79.53          & 87.63          ~/~ 65.49          & 1.97          & 4.36          & 9.49           & 33.84          & 68.51          \\
&SVHN     & 60.00          & 88.25          & 81.46          ~/~ 93.63          & 14.90         & 25.50         & 38.79          & 56.46          & 68.51          \\
& CIFAR-10    & 83.35          & 76.18          & 78.92          ~/~ 71.15          & 1.99          & 5.58          & 17.27          & 42.11          & 68.51          \\
& Tiny-ImageNet & 81.73          & 77.18          & 86.00          ~/~ 61.67          & 0.67          & 4.82          & 17.80          & 41.72          & 61.80          \\
& LSUN &  78.70          & 76.79          & 84.74          ~/~ 63.05          & 1.59          & 5.34          & 18.04          & 44.70          & 67.10          \\
& Places365 &  73.86          & 79.87          & 65.36          ~/~ 89.60          & 1.96          & 6.33          & 22.03          & 47.97          & 69.83          \\
\cmidrule{2-10}
& \textbf{Mean} & \textbf{75.45} & \textbf{79.63} & \textbf{80.69} ~/~ \textbf{74.10} & \textbf{3.85} & \textbf{8.66} & \textbf{20.57} & \textbf{44.47} & \textbf{67.38} \\
\bottomrule
\end{tabular}
\end{table*}

\begin{table*}[btp!]
\setlength{\tabcolsep}{6pt}
\caption{\textbf{Performance details on CIFAR-10 benchmark using WideResNet-28.}
UDG obtains consistently better results across OOD detection metrics. Accuracy shows the classification accuracy on all the (filtered) ID test samples.}
\label{T:appendix_cifar10_densenet}
\centering
\begin{tabular}{c|c|ccc|cccc|c}
\toprule
\multirow{2}{*}{Method} & \multirow{2}{*}{Dataset} 
& \multirow{2}{*}{FPR95~$\downarrow$} 
& \multirow{2}{*}{AUROC~$\uparrow$} 
& \multirow{2}{*}{AUPR(In/Out)~$\uparrow$}
& \multicolumn{4}{c|}{CCR@FPR~$\uparrow$} 
& \multirow{2}{*}{Accuracy~$\uparrow$} \\ 
\cmidrule(lr){6-9}
& && && $10^{-4}$  & $10^{-3}$  & $10^{-2}$ & $10^{-1}$ \\ 
\midrule
\multirow{7}{*}{\begin{tabular}[c]{@{}c@{}}MSP\end{tabular}}
&Texture     & 50.16          & 89.68          & 92.45          ~/~ 81.81          & 0.00          & 0.04          & 12.16          & 76.32          & 96.08          \\
&SVHN     & 30.54          & 95.44          & 92.81          ~/~ 97.49          & 8.75          & 25.94         & 72.94          & 89.16          & 96.08          \\
& CIFAR-100         & 51.38          & 89.15          & 87.42          ~/~ 87.99          & 0.02          & 0.77          & 11.15          & 75.25          & 96.08          \\
& Tiny-ImageNet & 56.98          & 88.96          & 90.14          ~/~ 84.19          & 0.03          & 0.71          & 13.85          & 75.72          & 93.69          \\
& LSUN &  47.05          & 90.54          & 88.99          ~/~ 89.44          & 0.20          & 0.80          & 11.97          & 79.25          & 96.08          \\
& Places365 &  53.44          & 89.18          & 70.65          ~/~ 95.54          & 0.04          & 0.74          & 9.22           & 75.86          & 95.02          \\
\cmidrule{2-10}
& \textbf{Mean} & \textbf{48.26} & \textbf{90.49} & \textbf{87.08} ~/~ \textbf{89.41} & \textbf{1.51} & \textbf{4.83} & \textbf{21.88} & \textbf{78.59} & \textbf{95.51} \\
\midrule
\multirow{7}{*}{\begin{tabular}[c]{@{}c@{}}ODIN\end{tabular}}
&Texture     & 47.50          & 81.23          & 82.94          ~/~ 78.25          & 0.00          & 0.00          & 1.81          & 32.69          & 96.08          \\
&SVHN     & 51.17          & 85.36          & 68.02          ~/~ 93.53          & 1.10          & 3.54          & 13.08         & 53.04          & 96.08          \\
& CIFAR-100         & 52.92          & 79.47          & 73.57          ~/~ 82.59          & 0.00          & 0.36          & 3.97          & 30.55          & 96.08          \\
& Tiny-ImageNet & 54.86          & 80.39          & 78.82          ~/~ 79.48          & 0.01          & 0.36          & 3.12          & 33.69          & 93.69          \\
& LSUN &  46.53          & 81.86          & 75.70          ~/~ 85.03          & 0.25          & 0.68          & 3.91          & 33.49          & 96.08          \\
& Places365 &  49.03          & 81.49          & 49.84          ~/~ 93.60          & 0.04          & 0.55          & 3.72          & 33.14          & 95.02          \\
\cmidrule{2-10}
& \textbf{Mean} & \textbf{50.33} & \textbf{81.63} & \textbf{71.48} ~/~ \textbf{85.41} & \textbf{0.23} & \textbf{0.91} & \textbf{4.94} & \textbf{36.10} & \textbf{95.51} \\
\midrule
\multirow{7}{*}{\begin{tabular}[c]{@{}c@{}}EBO\end{tabular}}
&Texture     & 40.44          & 89.55          & 91.16          ~/~ 84.41          & 0.00          & 0.00          & 5.41           & 71.35          & 96.08          \\
&SVHN     & 16.13          & 96.90          & 93.77          ~/~ 98.47          & 2.93          & 18.26         & 68.48          & 91.28          & 96.08          \\
& CIFAR-100         & 42.41          & 88.97          & 85.73          ~/~ 89.42          & 0.01          & 0.72          & 8.77           & 67.94          & 96.08          \\
& Tiny-ImageNet & 45.81          & 89.55          & 89.55          ~/~ 86.72          & 0.03          & 0.61          & 9.93           & 73.79          & 93.69          \\
& LSUN &  37.14          & 90.58          & 87.47          ~/~ 91.07          & 0.29          & 0.83          & 8.51           & 76.21          & 96.08          \\
& Places365 &  39.84          & 89.86          & 68.32          ~/~ 96.33          & 0.04          & 0.68          & 7.15           & 73.24          & 95.02          \\
\cmidrule{2-10}
& \textbf{Mean} & \textbf{36.96} & \textbf{90.90} & \textbf{86.00} ~/~ \textbf{91.07} & \textbf{0.55} & \textbf{3.52} & \textbf{18.04} & \textbf{75.64} & \textbf{95.51} \\
\midrule
\multirow{7}{*}{\begin{tabular}[c]{@{}c@{}}MCD\end{tabular}}
&Texture     & 93.19          & 70.58          & 82.49          ~/~ 49.12          & 0.00             & 0.15          & 7.65           & 44.96         & 87.85          \\
&SVHN     & 88.68          & 81.37          & 74.43          ~/~ 86.75          & 3.28          & 8.65          & 28.28          & 66.86         & 87.85          \\
& CIFAR-100         & 83.29          & 76.58          & 77.17          ~/~ 72.50           & 0.03          & 0.72          & 10.47          & 45.36         & 87.85          \\
& Tiny-ImageNet & 86.6           & 74.83          & 80.53          ~/~ 64.30           & 0.04          & 2.48          & 12.88          & 44.47         & 85.58          \\
& LSUN &  93.06          & 70.14          & 72.62          ~/~ 63.38          & 0.55          & 2.81          & 10.51          & 36.16         & 87.85          \\
& Places365 &  93.13          & 70.42          & 49.04          ~/~ 84.32          & 0.10           & 2.39          & 9.65           & 36.37         & 86.48          \\
\cmidrule{2-10}
& \textbf{Mean} & \textbf{89.66} & \textbf{73.99} & \textbf{72.71} ~/~ \textbf{70.06} & \textbf{0.67} & \textbf{2.87} & \textbf{13.24} & \textbf{45.7} & \textbf{87.24} \\
\midrule
\multirow{7}{*}{\begin{tabular}[c]{@{}c@{}}OE\end{tabular}}
&Texture     & 35.14          & 92.44          & 95.27          ~/~ 87.17          & 5.27          & 8.94           & 31.17          & 79.23          & 94.95          \\
&SVHN     & 22.94          & 96.23          & 94.14          ~/~ 97.78          & 37.34         & 52.79          & 73.87          & 88.74          & 94.95          \\
& CIFAR-100         & 52.99          & 87.17          & 86.80          ~/~ 86.09          & 1.72          & 6.83           & 21.22          & 63.16          & 94.95          \\
& Tiny-ImageNet & 55.53          & 87.43          & 90.20          ~/~ 82.58          & 4.58          & 13.91          & 28.61          & 64.92          & 92.72          \\
& LSUN &  59.69          & 85.56          & 86.18          ~/~ 83.67          & 5.18          & 11.55          & 26.09          & 58.88          & 94.95          \\
& Places365 &  55.30          & 85.75          & 69.15          ~/~ 94.25          & 4.50          & 10.31          & 22.42          & 56.79          & 94.24          \\
\cmidrule{2-10}
& \textbf{Mean} & \textbf{46.93} & \textbf{89.10} & \textbf{86.96} ~/~ \textbf{88.59} & \textbf{9.76} & \textbf{17.39} & \textbf{33.90} & \textbf{68.62} & \textbf{94.46} \\
\midrule
\multirow{7}{*}{\begin{tabular}[c]{@{}c@{}}UDG\end{tabular}}
&Texture     & 22.59          & 95.86          & 97.49          ~/~ 92.59          & 0.87           & 8.92           & 58.06          & 87.56          & 94.50          \\
&SVHN     & 17.23          & 97.23          & 95.43          ~/~ 98.64          & 45.32          & 60.75          & 78.46          & 89.84          & 94.50          \\
& CIFAR-100         & 43.36          & 91.53          & 92.08          ~/~ 90.21          & 5.19           & 12.28          & 37.79          & 77.03          & 94.50          \\
& Tiny-ImageNet & 39.33          & 93.90           & 95.90           ~/~ 90.01          & 4.86           & 27.52          & 64.17          & 82.97          & 92.07         \\
& LSUN &  30.17          & 95.25          & 96.06          ~/~ 94.05          & 13.28          & 36.98          & 66.03          & 86.35          & 94.50          \\
& Places365 &  35.24          & 94.31          & 89.24          ~/~ 97.55          & 8.39           & 27.67          & 61.10           & 83.75          & 93.33         \\
\cmidrule{2-10}
& \textbf{Mean} & \textbf{31.32} & \textbf{94.68} & \textbf{94.36} ~/~ \textbf{93.84} & \textbf{12.98} & \textbf{29.02} & \textbf{60.93} & \textbf{84.58} & \textbf{93.90} \\
\bottomrule
\end{tabular}
\end{table*}

\begin{table*}[btp!]
\setlength{\tabcolsep}{6pt}
\caption{\textbf{Performance details on CIFAR-100 benchmark using WideResNet-28.}
UDG obtains consistently better results across OOD detection metrics. Accuracy shows the classification accuracy on all the (filtered) ID test samples, which can be improved by UDG on the top of OE method.}
\label{T:appendix_cifar100_densenet}
\centering
\begin{tabular}{c|c|ccc|cccc|c}
\toprule
\multirow{2}{*}{Method} & \multirow{2}{*}{Dataset} 
& \multirow{2}{*}{FPR95~$\downarrow$} 
& \multirow{2}{*}{AUROC~$\uparrow$} 
& \multirow{2}{*}{AUPR(In/Out)~$\uparrow$}
& \multicolumn{4}{c|}{CCR@FPR~$\uparrow$} 
& \multirow{2}{*}{Accuracy~$\uparrow$} \\ 
\cmidrule(lr){6-9}
& && && $10^{-4}$  & $10^{-3}$  & $10^{-2}$ & $10^{-1}$ \\ 
\midrule
\multirow{7}{*}{\begin{tabular}[c]{@{}c@{}}MSP\end{tabular}}
&Texture     & 84.24          & 76.10          & 85.25          ~/~ 58.36          & 0.24          & 2.19          & 9.78           & 46.20          & 80.25          \\
&SVHN     & 79.63          & 78.95          & 65.45          ~/~ 88.22          & 1.42          & 4.26          & 17.14          & 51.39          & 80.25          \\
& CIFAR-10    & 77.07          & 80.81          & 83.16          ~/~ 76.76          & 0.49          & 9.19          & 25.03          & 53.94          & 80.25          \\
& Tiny-ImageNet & 81.25          & 79.12          & 87.75          ~/~ 63.33          & 0.31          & 5.34          & 24.75          & 51.64          & 72.92          \\
& LSUN &  81.32          & 78.51          & 86.81          ~/~ 62.95          & 0.51          & 2.57          & 20.03          & 50.74          & 78.54          \\
& Places365 &  75.28          & 80.84          & 67.81          ~/~ 89.76          & 1.49          & 4.63          & 20.12          & 53.24          & 80.03          \\
\cmidrule{2-10}
& \textbf{Mean} & \textbf{79.80} & \textbf{79.05} & \textbf{79.37} ~/~ \textbf{73.23} & \textbf{0.74} & \textbf{4.70} & \textbf{19.48} & \textbf{51.19} & \textbf{78.71} \\
\midrule
\multirow{7}{*}{\begin{tabular}[c]{@{}c@{}}ODIN\end{tabular}}
&Texture     & 78.88          & 76.46          & 84.68          ~/~ 62.45          & 0.15          & 1.52          & 10.21          & 41.44          & 80.25          \\
&SVHN     & 92.26          & 68.41          & 49.07          ~/~ 81.28          & 1.73          & 2.93          & 8.02           & 28.93          & 80.25          \\
& CIFAR-10    & 78.22          & 80.14          & 81.43          ~/~ 76.26          & 0.06          & 3.09          & 15.78          & 50.75          & 80.25          \\
& Tiny-ImageNet & 80.54          & 77.88          & 85.89          ~/~ 62.67          & 0.24          & 2.25          & 13.97          & 45.53          & 72.92          \\
& LSUN &  78.11          & 78.66          & 85.57          ~/~ 65.68          & 0.19          & 1.26          & 11.69          & 45.32          & 78.54          \\
& Places365 &  73.62          & 80.57          & 63.79          ~/~ 90.13          & 0.86          & 2.79          & 13.03          & 47.47          & 80.03          \\
\cmidrule{2-10}
& \textbf{Mean} & \textbf{80.27} & \textbf{77.02} & \textbf{75.07} ~/~ \textbf{73.08} & \textbf{0.54} & \textbf{2.31} & \textbf{12.12} & \textbf{43.24} & \textbf{78.71} \\
\midrule
\multirow{7}{*}{\begin{tabular}[c]{@{}c@{}}EBO\end{tabular}}
&Texture     & 84.22          & 76.13          & 85.08          ~/~ 58.51          & 0.08          & 1.55          & 10.04          & 44.24          & 80.25          \\
&SVHN     & 80.05          & 79.88          & 65.44          ~/~ 88.37          & 0.97          & 3.88          & 14.93          & 50.85          & 80.25          \\
& CIFAR-10    & 76.18          & 81.50          & 83.34          ~/~ 77.36          & 0.45          & 6.11          & 21.03          & 53.73          & 80.25          \\
& Tiny-ImageNet & 80.78          & 79.94          & 88.02          ~/~ 64.18          & 0.06          & 4.92          & 22.31          & 51.82          & 72.92          \\
& LSUN &  82.59          & 78.74          & 86.71          ~/~ 62.94          & 0.64          & 1.55          & 17.71          & 49.76          & 78.54          \\
& Places365 &  74.54          & 81.63          & 67.67          ~/~ 90.18          & 1.13          & 3.69          & 17.55          & 52.47          & 80.03          \\
\cmidrule{2-10}
& \textbf{Mean} & \textbf{79.73} & \textbf{79.64} & \textbf{79.38} ~/~ \textbf{73.59} & \textbf{0.55} & \textbf{3.62} & \textbf{17.26} & \textbf{50.48} & \textbf{78.71} \\
\midrule
\multirow{7}{*}{\begin{tabular}[c]{@{}c@{}}MCD\end{tabular}}
&Texture     & 91.33          & 69.03          & 79.60          ~/~ 49.66          & 0.00          & 0.29          & 4.49           & 32.61          & 68.80          \\
&SVHN     & 87.03          & 73.48          & 52.89          ~/~ 84.73          & 1.74          & 2.90          & 6.68           & 33.88          & 68.80          \\
& CIFAR-10    & 86.89          & 73.79          & 76.15          ~/~ 68.38          & 0.26          & 2.88          & 13.40          & 39.94          & 68.80          \\
& Tiny-ImageNet & 85.16          & 74.59          & 84.19          ~/~ 58.36          & 1.01          & 2.58          & 13.71          & 40.31          & 62.22          \\
& LSUN &  88.67          & 72.04          & 83.06          ~/~ 54.33          & 1.13          & 3.58          & 15.95          & 39.58          & 67.29          \\
& Places365 &  86.83          & 74.05          & 59.58          ~/~ 85.28          & 1.24          & 3.66          & 14.85          & 41.07          & 69.77          \\
\cmidrule{2-10}
& \textbf{Mean} & \textbf{87.65} & \textbf{72.83} & \textbf{72.58} ~/~ \textbf{66.79} & \textbf{0.90} & \textbf{2.65} & \textbf{11.51} & \textbf{37.90} & \textbf{67.61} \\
\midrule
\multirow{7}{*}{\begin{tabular}[c]{@{}c@{}}OE\end{tabular}}
&Texture     & 93.07          & 67.00             & 78.92     ~/~ 46.52          & 0.02         & 0.52          & 5.50            & 32.16         & 74.01         \\
&SVHN     & 88.74          & 76.14          & 66.07          ~/~ 85.17          & 7.06         & 12.91         & 24.82          & 47.43         & 74.01         \\
& CIFAR-10    & 78.82          & 79.36          & 81.29          ~/~ 75.27          & 1.08         & 7.63          & 17.49          & 48.84         & 74.01         \\
& Tiny-ImageNet & 83.34          & 78.35          & 87.34          ~/~ 61.78          & 1.06         & 8.84          & 24.40           & 47.64         & 66.49         \\
& LSUN &  84.96          & 78.11          & 87.26          ~/~ 60.76          & 5.80          & 10.40          & 25.75          & 48.27         & 71.47         \\
& Places365 &  80.30           & 79.87          & 67.23          ~/~ 88.65          & 1.78         & 6.29          & 19.78          & 49.84         & 74.39         \\
\cmidrule{2-10}
& \textbf{Mean} & \textbf{84.87} & \textbf{76.47} & \textbf{78.02} ~/~ \textbf{69.69} & \textbf{2.80} & \textbf{7.76} & \textbf{19.63} & \textbf{45.70} & \textbf{72.40} \\
\midrule
\multirow{7}{*}{\begin{tabular}[c]{@{}c@{}}UDG\end{tabular}}
&Texture     & 73.62          & 79.01          & 85.53          ~/~ 67.08          & 0.00             & 0.00             & 6.74           & 46.09         & 75.77          \\
&SVHN     & 66.76          & 85.29          & 76.14          ~/~ 92.33          & 8.00             & 15.83         & 32.57          & 58.05         & 75.77          \\
& CIFAR-10    & 82.35          & 76.67          & 78.52          ~/~ 72.63          & 0.51          & 3.90           & 15.29          & 44.79         & 75.77          \\
& Tiny-ImageNet & 78.91          & 79.04          & 87.00             ~/~ 65.06          & 0.12          & 2.86          & 19.13          & 47.50          & 68.57          \\
& LSUN &  77.04          & 79.79          & 87.49          ~/~ 66.93          & 2.51          & 6.01          & 22.33          & 49.14         & 73.93          \\
& Places365 &  72.25          & 81.49          & 66.72          ~/~ 90.65          & 1.19          & 3.28          & 17.59          & 50.82         & 76.10           \\
\cmidrule{2-10}
& \textbf{Mean} & \textbf{75.16} & \textbf{80.21} & \textbf{80.23} ~/~ \textbf{75.78} & \textbf{2.05} & \textbf{5.31} & \textbf{18.94} & \textbf{49.40} & \textbf{74.32} \\
\bottomrule
\end{tabular}
\end{table*}
\end{document}